\documentclass[12pt,rapport]{dms}

\usepackage[utf8]{inputenc} 
\usepackage[T1]{fontenc}    %
\usepackage[table,xcdraw]{xcolor}
\usepackage{tabularx}
\usepackage{url}

\usepackage{lmodern}
\DeclareSymbolFont{largesymbols}{OMX}{cmex}{m}{n}

\anglais

\def\sloppy{%
  \tolerance 500
  \emergencystretch 3em%
  \hfuzz .5pt
  \vfuzz\hfuzz}
\usepackage{graphicx,amssymb,subfigure,icomma}

\usepackage[labelfont=bf, width=\linewidth]{caption}

\usepackage{hyperref}  
\hypersetup{colorlinks=true,allcolors=black}
\usepackage{hypcap}   
\usepackage{bookmark}
\usepackage[inkscapeformat=png]{svg}

\theoremstyle{definition}

\numberwithin{equation}{section}
\numberwithin{table}{chapter}
\numberwithin{figure}{chapter}

\begin{document}

\version{1}
\pagenumbering{roman}



\title{Building a privacy-preserving Federated Recommender system for mobile devices}

\author{Aasheesh Singh}

\copyrightyear{2024}

\department{Département d’informatique et de recherche opérationnelle}


\sujet{Apprentissage automatique}










\pdfbookmark[chapter]{Couverture}{PageUn}

\maketitle






\anglais
\chapter*{Abstract}
 
Traditionally, mobile applications have relied heavily on third party trackers and app usage data to serve recommendations on their platforms. In this paradigm, private user data is pooled into a centralized server and serves as a repository for training machine learning algorithms. Introduction of privacy regulations such as General Data Protection Regulation (GDPR) and a heightened sense of awareness amongst users regarding personal data, make it imperative to find new ways to serve relevant content to mobile users. Towards this end, in this work we investigate methods and build systems to develop privacy preserving hyper-personalized recommendation systems using Federated learning.\\ 

Federated learning is a decentralized approach to training machine learning models where instead of moving all data to a central database, the model is trained locally on each device and only the model updates are shared without any data sharing. In this work, we build a two stage pipeline for a federated recommendation system where the first stage involves candidate generation of relevant items and the second stage involves ranking of these items hyper-personalized to a given user. The second stage of the pipeline generally involves collecting sensitive mobile user data such as demographics, location, mobile usage habits, sensor data etc. This stage of the pipeline is federated in our proposed system such that model improvements can be done in a private fashion without actual data leaving the user device. \\

We employ various open-source datasets such as MovieLens \cite{movielens_dataset}, Human Activity Recognition datasets \cite{UCI_HAR} and a small pilot dataset by the host company Lerna AI for the subject of this work. The goal of this work was to build a proof-of-concept system for a Federated recommendation engine that can be deployed for a beta-testing pilot on mobile devices. As such, equal amount of efforts have been dedicated towards research and engineering deployment. The final system is a machine learning library on Kotlin Multiplatform \cite{jetbrains_kotlin_multiplatform} that is deployed on Android/iOS user devices.

\anglais
\cleardoublepage
\pdfbookmark[chapter]{\contentsname}{toc}  
\tableofcontents
\cleardoublepage
\phantomsection  
\listoftables
\cleardoublepage
\phantomsection
\listoffigures


\chapter*{List of acronyms and abbreviations}
\begin{twocolumnlist}{.2\textwidth}{.7\textwidth}
  GDPR & General Data Protection Regulation\\
  CTR & Click-through-rate\\
  SDK & Software Development Kit\\
  HAR & Human Activity Recognition\\
  UCI & University of California, Irvine\\
  ADL & Activities of Daily Living\\
  t-SNE & t-Distributed Stochastic Neighbor Embedding\\
  PacMAP & Pairwise Controlled Manifold Approximation\\
  CCO & Correlated Cross-Occurrence\\
  NCF & Neural Collaborative Filtering\\
\end{twocolumnlist}


\chapter*{Acknowledgements}

I would like to express my sincere gratitude to the host company Lerna AI and MITACS for providing me the opportunity to undertake this internship as part of my Master’s degree. I am grateful to my supervisors at Lerna AI, Dr Georgios Kellaris and Georgios Depastas for their exceptional guidance, support, and constructive feedback throughout the internship. I particularly appreciate the work environment and flexibility provided by the company during the course of this internship with full-time and part-time components. I deeply enjoyed working on both research and engineering challenges encountered during the course of this internship with the Lerna AI team.   \\

I would also like to thank the internship coordination teams at MILA-Quebec AI Institute and Université de Montréal for smooth facilitation and approval of the internship process. I am grateful to them for providing the necessary resources and environment for me to pursue my Master's degree and complete this internship. Further, I want to extend my heartfelt gratitude to my mentor, Fuyuan Lyu, Phd student at MILA for his research guidance during our weekly meetings. His feedback and support in finding relevant literature immensely helped in steering the course of the internship. \\

Finally, I want to thank my family: my sister, brother-in-law and my mother whose unwavering support allowed me to pursue my academic goals. Their patience, love and understanding enabled me to push above my abilities. I could not have achieved this academic milestone in my life without their support.

 %
 %

\cleardoublepage
\pagenumbering{arabic}





\chapter*{Introduction}


\textbf{Problem Context}
\newline

Personalization of software products and services is one of the most important drivers for a positive customer experience. Hyper-personalization by offering recommendations instead of fixed offerings reduces user churn and increases conversion rate. Increasingly, companies are focusing their services for mobile-first usage as opposed to traditional web browsing as smartphones have become ubiquitous for accessing internet and consuming content. \\ 

The two key decisions towards hyper-personalization in mobile-first applications are: “when” to send notifications to a user and “what” content to show. The first task relating to the “when” part, nudges users with personalized notifications according to their usage habits when they would be most likely to engage. The second task relating to the “what” part, involves recommending curated content from a database of items which are most relevant based on user interests. A simple example of this pipeline would be a mobile app which curates movie recommendations at dinner time or short-form content for lunch time by learning user habits \& interests.\\

\textbf{Solution}
\newline

In this work, we tackle the second part of this pipeline i.e. what content to show that is personalized to a user. A recommendation engine typically ingests either or both of these two categories of data: user features and item features. User features consists of demographics data (age, gender, city etc.) and interaction history (purchases, clicks, views etc.), whereas item features correspond to the metadata associated with the product. For mobile applications, another stream of user features is available in terms of sensor data, location, usage habits, device state etc. This data is sensitive in nature and holds immense potential for hyper-personalization but traditional recommendation systems fail to address the privacy concerns of users to utilize this in a secure and private fashion. Towards this end, we build a system for serving recommendations to mobile users in a federated fashion where their sensitive data never leaves the device. \\

The proposed system in this work is a two-stage pipeline where the first stage is a collaborative filtering method based on a cross co-occurrence algorithm \cite{ferrel_unified_recommender} which generates an initial candidate pool of relevant items, while the second stage is a Click-through-rate(CTR) prediction algorithm that generates the list of top-k items which are hyper-personalized to a given user from the given candidate pool. The latter stage involves private user data and is thus a federated machine learning system whereas the first stage is a centralized system deployed on cloud servers.\\

\textbf{Project Objectives Division}
\newline

This work is divided into chapters, each describing the sub-task being solved for creating the system. Each of the sub-tasks are largely independent problems towards the full pipeline and the report is organized such that each chapter is self-contained with its own literature review, dataset description, methodology and results discussion. The chapter distribution is as following:\\

\begin{enumerate}
    \item \textbf{Chapter-1}: Company introduction and Project scope description, timeline of work, software used and tooling.\\
    \item \textbf{Chapter-2}: Harnessing data from Mobile devices. In this chapter we list the complete suite of sensor and device data collected from user mobile devices which is used for personalization. We detail our methods for extracting useful information from raw sensor data such as accelerometers, gyroscope etc for the problem of Activity recognition in Part-A of this chapter. In part-B of this chapter, we detail our experiments for a pilot study dataset of the company and demonstrate how solving the Activity recognition problem improves efficacy in this task \\
    \item \textbf{Chapter-3}:Two-stage Recommendation System. In this chapter, we detail our proposed approach for hyper-personalization using a two-stage Recommendation pipeline. The first stage is a centralized stage for Candidates generation of items from a large database and is detailed in this chapter.\\
    \item \textbf{Chapter-4}: Federated Ranking System. In this chapter, we detail our approach for the second stage of our pipeline which involves sensitive user data from mobile devices. This step is the federated learning stage of our pipeline and is deployed on user mobile devices.\\
    \item \textbf{Chapter-5}: Conclusion. We finally conclude our report listing the Key contributions of our work during this internship, the main challenges encountered and further future work directions.
\end{enumerate}

\chapter{Company and Project Scope}

\section{Company Description: Lerna AI}

Lerna AI is a SaaS company that helps mobile apps boost their notification and product recommendation conversions, with their mobile hyper-personalization recommender. Their mobile SDK available for multi-platform devices (Android/iOS) enables apps to personalize their content to each user, by training models on content metadata along with user data (demographic \& sensor), privacy-preserving on-device. Traditionally, apps have heavily depended on trackers and third-party data. Due to privacy laws and blockers, this is no longer possible. Lerna AI’s SDK library allows app publishers to collect real-time behavioural insights and prediction, based on enriched first-party data that never leaves that device, forgoing the need to rely on expensive 3rd party data. This is achieved with the use of cutting-edge privacy-preserving Federated Learning, differential privacy and Secure multi-party computation techniques. Such predictive capability, can help app publishers increase revenue, extend use time in-app, and reduce churn.\\

\begin{figure}[h]
    \centering
    \includegraphics[width=0.9\textwidth]{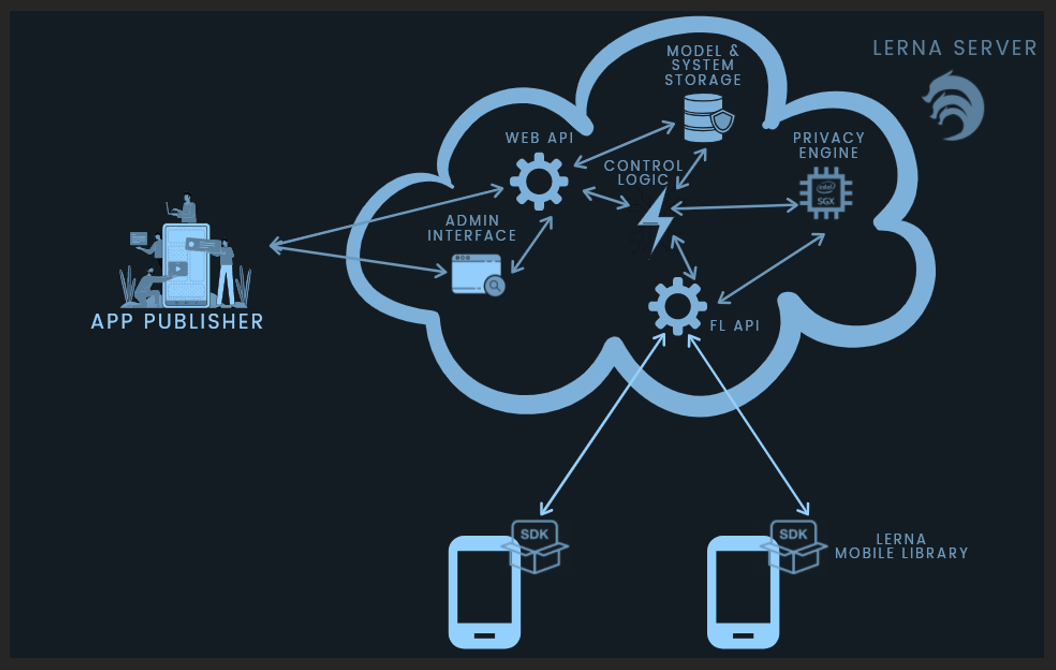}
    \caption{High level architecture of the company’s product offering which is designed for App owners or publishers. The company’s offering consists of a mobile SDK library which is integrated into the app code to provide federated recommendations. Further, a fully managed cloud server coordinates the model weight updates from edge devices with a differential privacy engine. Services such as dashboards and monitoring are also offered.}
    \label{fig:1.1}
\end{figure}

A high-level architecture of the company’s product offering is shown in the \textbf{Fig \ref{fig:1.1}}.\\

The company is an early-stage startup founded in 2020 and is based out of Montreal, Quebec. The co-founders: Georgios Depastas and Dr. Georgios Kellaris have deep expertise in building privacy and fintech products. The company consists of 3 members including the above founders and an engineer. The work mode for this internship was hybrid, consisting of both in-office and remote components.

\section{Project Scope}

\subsection{Terminology}

\textbf{Federated Learning}
\newline

Federated learning(\cite{federated_learning_def}) is a decentralized approach to training machine learning models. The traditional way of deploying machine learning on edge devices includes sending data from the devices to a central cloud server where it is aggregated and processed before training the model on the cloud. The updated model can then either be called through a REST API endpoint, or if it’s computationally viable, can be deployed for inference-only on the mobile device by downloading the updated weights from the cloud. \\

In the decentralized paradigm of federated learning however, the actual data from the client devices is never sent to the cloud server. Instead, the raw data on edge devices is used to train the model locally and the model updates are sent to the cloud where they are aggregated to form a global model. This increases data privacy and is increasingly gaining prominence in sectors where sensitive data is involved. \\

\textbf{Differential Privacy}
\newline

Differential privacy(\cite{differential_privacy}) is a mathematical framework for privacy which allows sharing information about a group of individuals, by describing statistical properties within the group while withholding information about specific individuals. In practical terms, differential privacy adds noise or randomness to data before sharing. This noise is carefully calibrated to ensure that statistical properties of the dataset are preserved while preventing the reconstruction of individual data points. For example, to calculate mean statistic of a population, the method would add a different noise value to each data point which collectively sums to zero in the final average of the calculation. Thus the privacy of each individual is preserved without any effect on the output.\\

In the context of federated learning, a noise value is added to each parameter of the model weights such that during weight aggregation in the central server the total noise sums to zero. Differential privacy therefore adds another layer of privacy on top of federated learning, such that even the model updates being shared are private and there is no possibility of retrieving any user data from the model weights by a third party attacker.\\

\section{Objectives}

\textbf{System Description}
\newline

The company’s existing system is a logistic regression model that concatenates item metadata features, user demographics features and sensitive data from mobile devices into one single input to predict a relevance score which is used for recommendation. This small model is deployed on edge mobile devices and the local weight updates after training are shared to a central server for federated machine learning. Before sending the weight updates to the cloud, the weights are added with a noise parameter to enforce differential privacy in the system. \\

The Logistic regression model and all ML training components on edge devices are implemented in Kotlin Multiplatform(\cite{jetbrains_kotlin_multiplatform}) library which communicates with the central server using REST API. The central server in the cloud runs a C service which coordinates the data sources and the model creator, and distributes tailored noise and secret shares to the parties in order to ensure privacy/security. Finally, the app developer embeds the Kotlin library into the application code to serve recommendations and monitors model performance using a web dashboard.\\

\begin{figure}[h]
    \centering
    \includegraphics[width=0.9\textwidth]{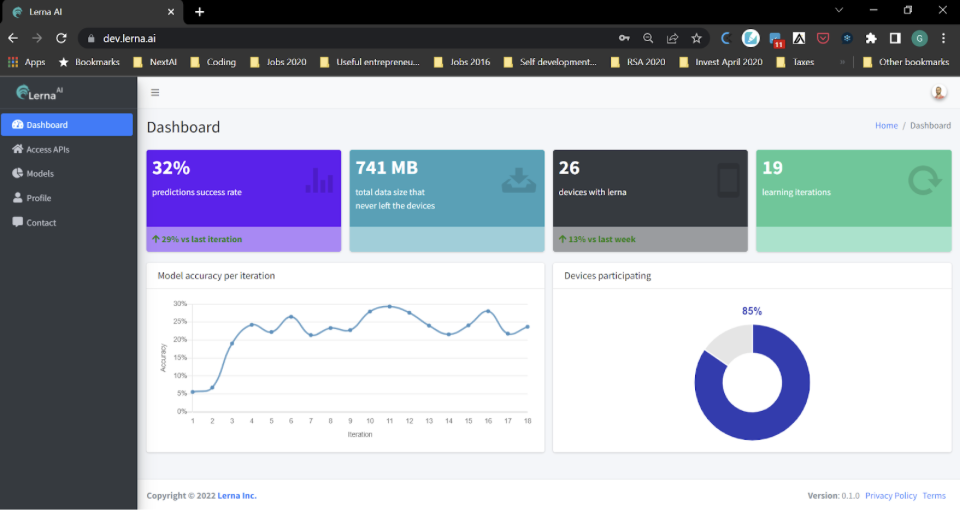}
    \caption{Lerna AI's Dashboard system, monitoring model performance and active mobile devices contributing to the Federated learning network.}
    \label{fig:1.2}
\end{figure}

\textbf{Project objectives}
\newline

The objectives of the internship were to improve upon the Logistic Regression model for delivering federated recommendations and implement corresponding algorithms from scratch in low-level Kotlin programming language for end-to-end deployment. The tasks performed in this internship are therefore broken down as following:\\

\begin{enumerate}
    \item Surveying existing literature, public datasets and initial problem formulation of the task.
    \item EDA of input sensor data sources from mobile device and transform raw data into meaningful abstractions such as activity labels or embeddings for usage in downstream task. This is covered in the Part-A subtask called Activity Recognition in Chapter-2 of this report.
    \item The efficacy of these sensor data transformations for the baseline Logistic regression model is explored on a pilot user dataset of the company. This is detailed in Part-B subtask, called Backtesting on a Pilot dataset in Chapter-2 of this report.
    \item Proposal of a two-stage novel pipeline for Federated recommendations. The two-stage pipeline involves a candidate generation and a ranking phase. The candidate generation phase is a centralized collaborative filtering algorithm which doesn’t utilize any sensitive mobile data. Chapter-3 of this report discusses the overall two-stage pipeline and further details the methodology of first centralized stage of the pipeline along with experimental results.
    \item The second stage of our pipeline is a Click-through-rate prediction algorithm which is deployed on user's mobile devices and is a federated system. Implementation of the algorithm and its centralized vs federated evaluation experiments are detailed in Chaper-4 of this report. 
    \item Finally, the end-to-end implementation of the proposed ML models from scratch in low level Kotlin Multiplatform is another key contribution of this work.\\
\end{enumerate}

\vspace{-8mm}
\section{Timeline}

The internship was completed in 8 months from May 08, 2023 to mid January, 2024 with a 4 month full-time component during Summer 2023 and a 4 months part-time component during the Fall 2023 academic semester. There was a period of brief two week break in the internship and was compensated with additional time spent later. Overall, the timeline of work for the tasks detailed in the previous section is as following:\\
\begin{enumerate}
    \item In May, significant effort was spent in finding public datasets and literature suitable for the task of federated recommendations. However, no such public dataset for recommendations was found that contained sensor data from user mobile devices along with product metadata and user demographics.  The problem was thus broken down into sub-tasks each with their own individual public datasets. 
    \item In June-July, the task of Activity Recognition from sensor data was investigated and its performance evaluated against a pilot user dataset collected by the company last year.
    \item In August-September, literature survey and methods for recommendation systems including collaborative filtering, CTR prediction, pairwise ranking algorithms were investigated on public datasets such as Movielens 1M \cite{movielens_dataset}, Retail Rocket \cite{roman_zykov_noskov_artem_anokhin_alexander_2022} and Yelp ratings\cite{yelp_dataset}. The proposed system of two-stage recommendations was explored.
    \item In Oct-Nov, experiments for the second stage of the pipeline for federated CTR prediction were performed on MovieLens 1M dataset. A centralized vs federated evaluation was performed using Flower framework(\cite{flower_ai}) in python.
    \item In Dec-January, the final ML model was implemented in Kotlin Multiplatform from scratch. Due to the absence of any Machine learning frameworks in Kotlin, the entire forward \& backward passes ML models, training, checkpointing and experimentation code was written from scratch. This is implemented as a standalone machine learning library in Kotlin Multiplatform and would be open-sourced soon. \\
\end{enumerate}

\section{Tooling and Work environment}

The main work machine used for the daily tasks was a Windows 10 system with WSL support for running linux subsystem. The training of machine learning models was performed on remote AWS EC2 Instances equipped with Nvidia T4 GPUs. PyCharm IDE was used as development environment for Python experiments and Android Studio for programming in Kotlin language.\\

PyTorch(\cite{pytorch}) was used for running ML experiments in Python and Flower framework (\cite{flower_ai}) was used for running federated learning experiments. Tensorboard was used for experiment tracking and libraries such as Matplotlib and Seaborn were used for visualization and plotting. Github was used for version control throughout the project.\\

The first-stage or our centralized recommendation system of the pipeline was deployed to production using the ActionML\cite{actionml} framework. The second-stage of the recommendation system was implemented as a new ML library for Kotlin Multiplatform.\\

\chapter{Harnessing Data from Mobile Devices}

\section{Introduction}

The core product of Lerna AI is a hyper-personalization SDK for mobile devices which provides a plug-and-play solution to app developers to offer personalized recommendations within their platform. The recommendations are powered by both item features, comprising product related information \& metadata, and user context features, including user preferences, location, demographics, and most importantly data originating from their mobile device. The last stream of data is particularly unique as it provides immediate context about the user's state and overall habits and whether the product or item being recommended is conducive to being clicked at the current moment. These combined features power a hyper-personalized recommendation system that delivers a unique product experience to each user, thereby boosting conversion rates for the company's clients, i.e. the app developers. 

Item features or product metadata are unique for each client, and their schema is defined by the app developer. In contrast, user context features are dictated by the Lerna AI SDK, which is embedded in the app on an end-user device. The core proposition of Lerna’s SDK is to extract valuable insights from sensitive user data in a privacy-preserving fashion, ensuring that this data never leaves the mobile device.
\newline

\textbf{Mobile Sensor Data}
\newline
 
The user context data is collected from the mobile device in background while the user is browsing a given application. This data consists of features which describe the context in which a user is interacting with the application. This includes raw sensor information such as readings from the gyroscope, accelerometer, magnetometer etc. which indicate whether the user is doing some activity or is browsing their phone at rest.  Further, device state information such as wifi/bluetooth connectivity indicate whether the user is at a familiar location such home, work or is at someplace new. An exhaustive \textbf{Table \ref{tab:sensors}} of the user-context features along with their description is discussed in \textbf{Section \ref{sec:2.2}}. For the task of Activity recognition however, we are only concerned with the 3 input sensor modalities namely: Linear Accelerometer, Gyroscope and Magnetometer. We discuss more about the task below. \\

\subsection{Part-A: Activity Recognition}

\begin{figure}[h]
    \centering
    \includegraphics[width=0.8\textwidth]{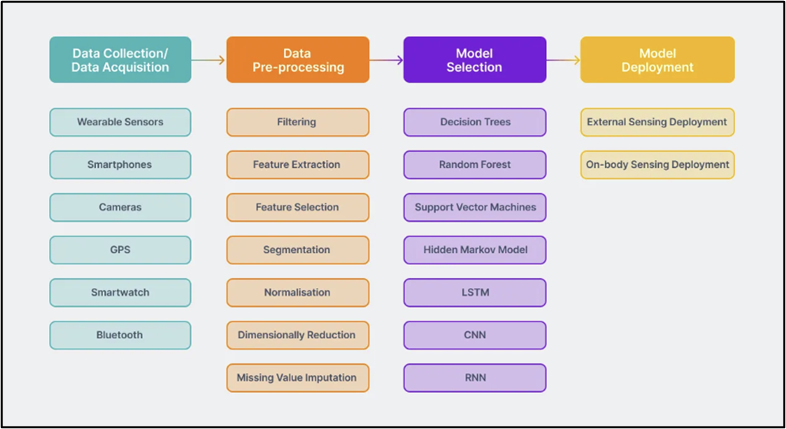}
    \caption{Human Activity Recognition pipeline\cite{1}.}
    \label{fig:2.1}
\end{figure}

To convert the raw sensor data from gyroscope, accelerometer etc. about motion and orientation into information that could be useful for downstream recommendation tasks, we experimented with methods that assign four discrete activity labels namely: Inactive, Active, Walking, Driving to a continuous sensor stream. These predicted classes could then be used as categorical one-hot features for a downstream task instead of working with time-series sensor data. Another approach experimented was to construct handcrafted features from signal processing literature which encode information from a fixed window of say 30 seconds of the sensor data. These handcrafted feature embeddings have higher representation capacity than raw sensor time-series data and allow usage of computationally simpler models for downstream tasks which are particularly useful for edge devices. In this chapter, we therefore discuss the methodology, datasets used and experiments for the task of Activity recognition from mobile sensor data and detail our findings which lay the groundwork for the rest of our pipeline. \textbf{Figure \ref{fig:2.1}} depicts the Activity Recognition pipeline.

\subsection{Part-B: Backtesting on a Pilot Dataset}
We wanted to test the hypothesis that processing raw time-series sensor data into either one-hot Activity labels or converting them into hand-crafted features provides a better input representation for the downstream recommendation models. We decided to conduct an experiment to test this hypothesis on a pilot study dataset about user’s mobile behaviour collected by the company last summer. The dataset was curated using data from mobile devices of 39 participants and had a dummy objective of predicting what application the user is going to open next based on their previous usage behaviour. It’s called a dummy objective as the participants were not asked to accomplish any task and their normal usage behaviour was recorded. 

Their performed actions were compared against the model predictions based on the context data recorded from their device listed in \textbf{Table \ref{tab:sensors}}.  The prior results on this dataset were based on a Logistic Regression model that inputs one big vector consisting of normalized sensor data and one-hot/multi-hot categorical features. Our goal was to conclude whether processing the raw continuous sensor data into class labels or feature embeddings improves performance compared to the previous results.

\section{Lerna SDK Sensor Suite}
\label{sec:2.2}
In this section we discuss the various input sources of data collected from the mobile device which contribute to forming the user-context features for the Lerna SDK. The data sources are clubbed into categories namely Sensors, Device Info, Connectivity, Audio etc. They are listed as following in \textbf{Table \ref{tab:sensors}}.\\

\renewcommand{\arraystretch}{1.42}
\begin{table}[ht]
\begin{tabular}{|c|c|c|}
\hline
\rowcolor[HTML]{FFFFFF} 
\textbf{Name}            & \textbf{Type of data} & \textbf{Domain}       \\ \hline
\rowcolor[HTML]{C2F4C0} 
Linear Accelerometer     & Sensors               & Continuous            \\ \hline
\rowcolor[HTML]{C2F4C0} 
Gyroscope                & Sensors               & Continuous            \\ \hline
\rowcolor[HTML]{C2F4C0} 
Magnetometer             & Sensors               & Continuous            \\ \hline
\rowcolor[HTML]{C2F4C0} 
Proximity Sensor         & Sensors               & Continuous            \\ \hline
\rowcolor[HTML]{C2F4C0} 
Ambient Light            & Sensors               & Continuous            \\ \hline
\rowcolor[HTML]{F4DEDD} 
Time of Day              & Date-Time             & Categorical           \\ \hline
\rowcolor[HTML]{F4DEDD} 
Type of Day              & Date-Time             & Categorical           \\ \hline
\rowcolor[HTML]{C9F6F4} 
Battery Percentage       & General Device Info   & Numeric               \\ \hline
\rowcolor[HTML]{C9F6F4} 
Screen Brightness        & General Device Info   & Numeric               \\ \hline
\rowcolor[HTML]{C9F6F4} 
Battery Charging         & General Device Info   & Categorical           \\ \hline
\rowcolor[HTML]{C9F6F4} 
Device Orientation       & General Device Info   & Categorical           \\ \hline
\rowcolor[HTML]{DAE8FC} 
Bluetooth Connections    & Connectivity          & Categorical           \\ \hline
\rowcolor[HTML]{DAE8FC} 
Wi-Fi Connection         & Connectivity          & Categorical           \\ \hline
\rowcolor[HTML]{FAE7D0} 
Volume Level             & Audio                 & Numeric               \\ \hline
\rowcolor[HTML]{FAE7D0} 
Audio Speaker On         & Audio                 & Categorical           \\ \hline
\rowcolor[HTML]{FAE7D0} 
Audio Headset On         & Audio                 & Categorical           \\ \hline
\rowcolor[HTML]{FAE7D0} 
Music Active             & Audio                 & Categorical           \\ \hline
\rowcolor[HTML]{FAE7D0} 
Wired Headset            & Audio                 & Categorical           \\ \hline
\rowcolor[HTML]{C9CBED} 
Previous App Duration    & App-related           & Numeric               \\ \hline
\rowcolor[HTML]{C9CBED} 
App Usage Categories     & App-related           & Multi-hot categorical \\ \hline
\rowcolor[HTML]{C9CBED} 
Installed App Categories & App-related           & Multi-hot categorical \\ \hline
\end{tabular}
\caption{The inputs from different categories of mobile sensors and device information which forms the input for the Lerna SDK}
\label{tab:sensors}
\end{table}

A brief description of the various data sources is as below:

\begin{itemize}
    \item \textbf{Linear accelerometer}: Continuous real-valued data capturing linear acceleration for each of the 3-dim axis (X,Y,Z).
    \item \textbf{Gyroscope}: Continuous sensor data capturing angular velocity along each of the 3 axis.
    \item \textbf{Magnetic field}: Continuous sensor data measuring the strength of earth’s magnetic field along the 3 axis.
    \item \textbf{Proximity sensor}: Continuous sensor data indicating the proximity of an object to the device.
    \item \textbf{Ambient light}: Continuous sensor data measuring the ambient light levels.
    \item \textbf{Time of Day}: Categorical data indicating the time of day. We discretize time into 4 intervals: Morning, Afternoon, Evening, Night.
    \item \textbf{Type of Day}: Categorical data indicating the type of day discretized into: weekdays or weekends.
    \item \textbf{Battery Percentage}: Numeric data indicating the current battery level percentage between 0-100.
    \item \textbf{Screen brightness}: Numeric data indicating the current screen brightness level. Range is (0-255).
    \item \textbf{Battery Charging}: Categorical data indicating whether the device is currently plugged in for charging or not.
    \item \textbf{Device Orientation}: Categorical data indicating the orientation of the device (portrait or landscape).
    \item \textbf{Bluetooth connections}: Categorical data indicating whether any Bluetooth devices are connected.
    \item \textbf{Wi-Fi connection}: Categorical data indicating whether the device is connected to a Wi-Fi network.
    \item \textbf{Volume Level}: Numeric data indicating the current volume level of the device.
    \item \textbf{Audio Speaker On}: Categorical data indicating whether the audio speaker is currently active and playing.
    \item \textbf{Audio Headset On}: Categorical data indicating whether a headset is currently connected and streaming audio.
    \item \textbf{Music Active}: Categorical data indicating whether music is currently playing.
    \item \textbf{Wired Headset}: Categorical data indicating whether a wired headset is currently connected.
    \item \textbf{Previous app duration}: Numeric data indicating the duration of the previous app usage in seconds.
    \item \textbf{App usage categories}: Multi-hot categorical data indicating the categories of apps currently in use. An app’s category is defined from the following categories: Games, News, Video(Netflix, YouTube), Audio (Spotify, Apple Music), Productivity(Drive, Calendar, Sheets), Delivery etc. The exhaustive list is discussed in the next Chapter.
    \item \textbf{Installed apps categories}: Multi-hot categorical data indicating the categories of installed apps.\\
\end{itemize}

\textbf{Sensor data used for Activity Recognition:}
\newline

Only the following sensors: Accelerometer, Gyroscope, Magnetometer, emit a continuous stream of time series data that changes rapidly during a short user activity session. The rest of the sensors, such as proximity, ambient light, device info, and connectivity, show very little variation during a typical user session, which typically lasts 10-60 seconds. Simple statistics such as mean (or mode for categorical variables) and standard deviation are typically sufficient to represent the changes in their state. Therefore, we are motivated to selectively address the time series data from these sensors to form meaningful higher order features, either as activity labels or as feature embeddings, which are the focus of this chapter. In the following section, we address the problem of activity recognition with reference to prior literature and explore public datasets for this task.

\section{Datasets}
We study public datasets from prior literature where sensor data from mobile devices are used for the task of Human Activity Recognition (HAR). Automatic recognition of human activity is valuable across wide range of application domains, primarily in sports and fitness tracking applications, healthcare monitoring,  elderly care, and increasingly in mobile recommendation systems.

Although numerous  public datasets for activity recognition are available such as \cite{mallol2021harage}, \cite{matey2023dataset}, they are primarily collected using wearable devices such as smartwatches which are not the target device for our problem statement. We were primarily interested in datasets that solve human activity recognition problem using just mobile devices which cater to a much larger user base. We found two public large scale HAR datasets that fit this requirement. We explore them in the sections below. 

\subsection{UCI Dataset}

\begin{figure}[h]
    \centering
    \includegraphics[width=0.9\textwidth]{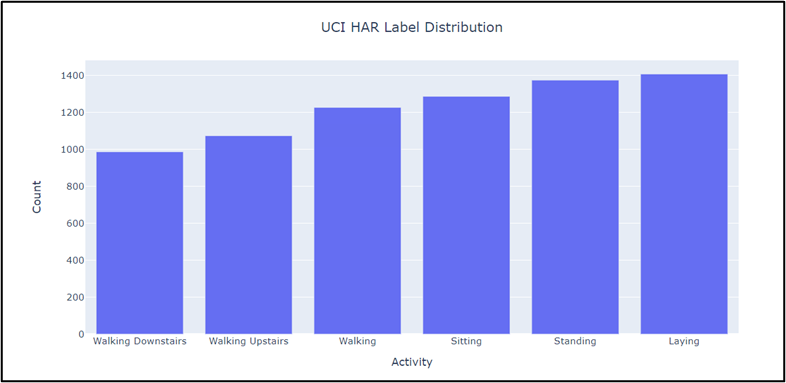}
    \caption{Data distribution for different Activity classes in UCI Dataset}
    \label{fig:2.2}
\end{figure}

This dataset curated by researchers from UC Irvine is publicly available on the UCI Machine Learning Repository \cite{UCI_HAR}. The dataset proposed in Anguita et al., 2013 \cite{UCI_HAR_PAPER} captures activities of daily living such as standing, sitting, lying down, walking etc. from a diverse group of 30 volunteers aged between 19 and 48 years. The authors of the dataset instructed the participants to follow a specific protocol of activities while wearing a Samsung Galaxy S II smartphone attached to their waist. \textbf{Figure \ref{fig:2.2}} depicts the data distribution for each of these classes.

\vspace{5mm}

\vspace{5mm}

The dataset captures time-series tri-axial acceleration data i.e. (tAcc-XYZ) from accelerometer, where "t" denotes time and the suffix "XYZ" denotes the tri-axial signal in X, Y and Z directions respectively. Additionally, tri-axial angular velocity data from a gyroscope sensor i.e., (tGyro-XYZ) was also recorded to understand rotation information. Both of these sensors were sampled at a frequency of 50Hz. The raw acceleration data recorded by the accelerometer was split into two components i.e. Body acceleration signals (tBodyAcc-XYZ) and Gravity acceleration (tGravityAcc-XYZ) using signal processing techniques. To separate the fixed gravity acceleration component from linear body acceleration, the authors used low-pass Butterworth filtering with a cutoff frequency of 20Hz. Along with this, median filtering was done on the output which resulted in separating the linear body acceleration component from total acceleration. This pre-processing is now used as a standard technique to remove correlated noise from the acceleration signal and improve downstream task performance. The continuous time-series data was then chunked into sessions of 2.56 seconds duration using a sliding windows with a 50\% overlap. More details about the dataset are detailed in \textbf{Table \ref{tab:differences}}.\\

\begin{figure}[h]
    \centering
    \includegraphics[width=0.9\textwidth]{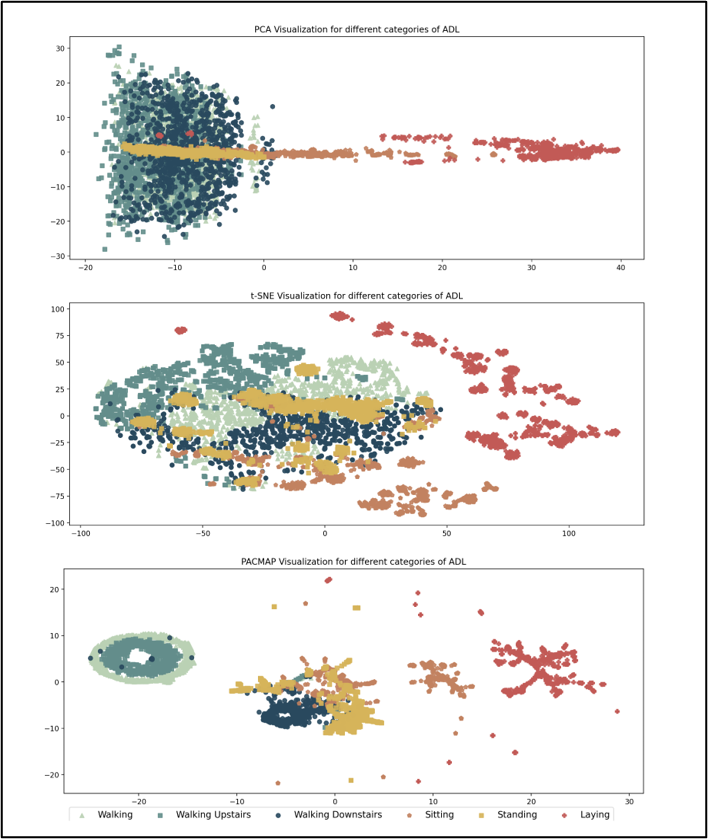}
    \caption{Dimensionality Reduction techniques to better understand UCI HAR Dataset}
    \label{fig:2.3}
    \vspace{-15mm}
\end{figure}

To understand the separation of various Activity classes, we leveraged various dimensionality reduction methods including Principal Component Analysis (PCA), t-Distributed Stochastic Neighbor Embedding (t-SNE), and Pairwise Controlled Manifold Approximation (PaCMAP) on the raw 6 dimensional input signals (3 axis acceleration, 3 gyroscope data.) The plots obtained are depicted in \textbf{Figure \ref{fig:2.3}}. PCA captures the global patterns by maximizing data variance along principal components, while t-SNE highlights the local patterns. PaCMAP\cite{PaCMAP} is a recent approach which captures both local and global structures and is comparatively faster than both. \textbf{Figure \ref{fig:2.3}} depicts that there is an evident non-linear separation among different meta-categories of Active and Inactive classes. The "Inactive Classes" such as Standing, Sitting, and Laying exhibit clear different clusters whereas for the "Active Classes" i.e., Walking, Walking Upstairs and Walking Downstairs, some overlap is evident between the clusters.

\vspace{2.5mm}
\subsection{Real-Life HAR Dataset}

The UCI dataset described above had been curated in a controlled laboratory setting with smartphones positioned around the waist. However, in real life scenarios individuals may use their smartphones in various orientations and placements, making it difficult for ML models to generalize across different users. Differences in height, smartphone placement across either hip or front pocket, different walking gait etc. are all factors which require a more free-from dataset without too many placement constraints. 

\vspace{4mm}
\renewcommand{\arraystretch}{0.99}
\begin{table}[h]
\begin{tabular}{lcc}
\hline
                                                & \textbf{UCI HAR}        & \textbf{Real Life HAR}     \\ \hline
\textbf{Types of action studied}                & Short-Themed            & Long-Themed               \\
\textbf{Smartphone orientation} & Fixed                   & Free                       \\
\textbf{Fixed sensor frequency}                 & Yes                     & No                         \\
\textbf{Sensors used}                           & Acc. and gyro.          & Acc., gyro., magn. and GPS \\
\textbf{No. of participants}                     & 30                      & 19                         \\
\textbf{Sliding Window}                        & 2.56s with 50\% overlap & 60s with 95\% overlap      \\ \hline
\end{tabular}
\caption{Differences between the UCI and Real Life HAR datasets}
\label{tab:differences}
\end{table}

\vspace{2.5mm}
To mitigate these challenges, Gonzalez et al.,2020 introduced the Real-life Human Activity Recognition dataset\cite{real_life_data}. The dataset was collected under realistic normal living scenarios, independent of orientation, placement, and subject variability. The dataset aims to provide a more trustworthy framework for activity recognition models in real-world applications. \textbf{Table \ref{tab:differences}} highlights the core differences between the two datasets: UCI dataset and Real-life HAR dataset based on their collection and preparation methodology. \\

The dataset consists of raw data from four activity monitoring sensors i.e., accelerometer, gyroscope, magnetometer, and GPS collected using a mobile application running on different Android devices. A group of 19 participants were selected to record activity sessions from their daily life habits and chores by using the mobile application to trigger and stop the collection of sensor data. Instead of having many activity labels relating to individual chores such as cooking, cleaning, studying etc., a few broad categories described below were chosen for the dataset.:\\

\begin{enumerate}
    \item \textbf{Inactive:} Not handling the mobile phone, such as leaving it on a desk while working or studying. 
    \item \textbf{Active:} Carrying the mobile phone while moving around, engaging in activities like cooking, attending events, shopping, or doing household chores.
    \item \textbf{Walking:} Moving to a specific location, including activities like running or jogging.
    \item \textbf{Driving:} Traveling in motorized vehicles such as cars, buses, or motorcycles.\\
\end{enumerate}

\begin{figure}[h]
    \centering
    \includegraphics[width=0.9\textwidth]{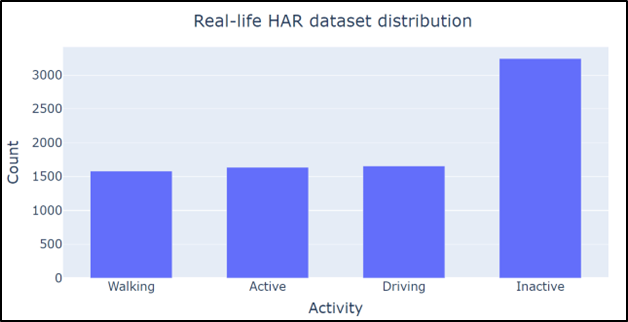}
    \caption{Data distribution for different classes in real life HAR dataset}
    \label{fig:2.4}
\end{figure}

\vspace{3mm}
\textbf{Figure \ref{fig:2.4}} denotes the class distribution for the above mentioned activities in the real-life HAR dataset. \\
\vspace{-5mm}

Acceleration, rotation, and magnetic field readings were recorded in tri-axial dimensions. GPS data encompassed latitude, longitude, altitude, bearing, speed, and accuracy metrics. Unlike the previous UCI dataset which used low-pass butterworth filter to obtain the body's linear acceleration data from raw acceleration values, this dataset utilized a separate dedicated accelerometer to offset gravity acceleration. The gravity accelerometer readings were subtracted from the raw acceleration recorded by the mobile devices to obtain the body's linear acceleration. This method provided clear linear acceleration readings that were unaffected by noise due to smartphone orientation. We only utilize the accelerometer and gyroscope sensor data from the dataset for our activity recognition experiments. In the original work, the continuous time-series sensor readings were divided into chunks or sessions of 20 seconds each, using sliding windows with a 19 second overlap (95\%) between two windows. However, for our experiments, we use 60 second chunks with same 95 \%  overlap to keep the session size consistent with our Lerna Pilot dataset. More details about the pre-processing are described in the Methodology section below. \\

\begin{figure}[h]
    \centering
    \includegraphics[width=0.9\textwidth]{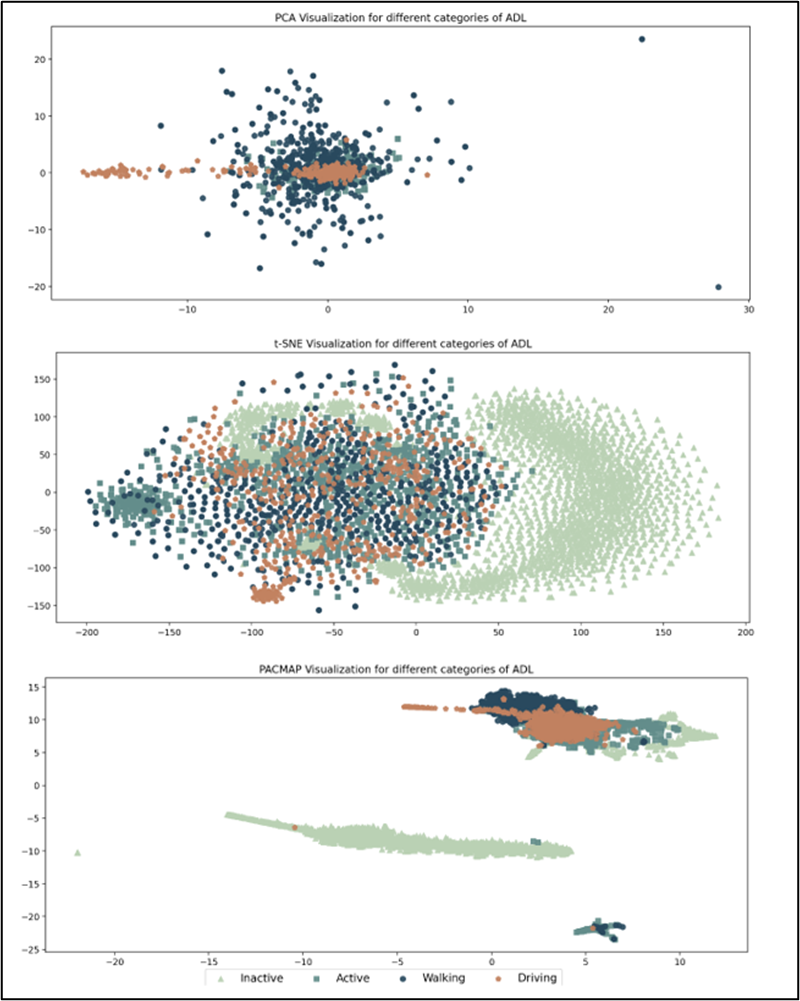}
    \caption{Dimensionality reduction techniques to better understand the Real life HAR dataset}
    \label{fig:2.5}
\end{figure}

Similar to the UCI dataset, we analysed the dataset using dimensionality reduction methods to understand the separation of various Activity classes as shown in \textbf{Figure \ref{fig:2.5}}. We can observe that the "Inactive" class has a clear distinct cluster in the t-SNE and PaCMAP plots. Moreover, "Driving" and "Walking" are also separate clusters as observed in the PCA and PaCMAP plot. However, there is no visibly clear cluster for the "Active" class due to it being a group class for multiple daily activities. \\

\textbf{UCI vs Real-life HAR dataset:}\\

Based on the exploration of both datasets, we chose Real-life HAR dataset as the primary dataset for our further experiments. The motivation being that it closely imitates real-world application scenarios of our system, and is independent of orientation and placement of mobile device. The 4 classes: Inactive, Active, Walking \& Driving provide the right level of granularity needed for understanding the device state for our mobile recommendation system pipeline.  \\

\subsection{Lerna Pilot Dataset}

Lerna AI last summer had launched the Hydra Demo app to test their mobile library integration in user devices. This dataset was collected as part of that pilot demo. The app predicts the next app category (defined in \textbf{Table \ref{tab:lerna_categories}} below) that a mobile user is more likely to open based on the context of sensor and usage data. The Hydra Demo app was not a commercial product for Lerna AI, but it incorporates the company’s underlying mobile library and allows Lerna AI to collect data (device info, app usage, sensor data etc.) that would be present in real world deployment of its recommendation library. 

The dataset consisted of 91 input features based on \textbf{Table \ref{tab:sensors}} and an output label denoting the next app category. The app labels are obtained from the app’s categorization in Android PlayStore. The output label app categories and data statistics are detailed below in \textbf{Table \ref{tab:lerna_data}} and \textbf{Table \ref{tab:lerna_categories}} below. The data is bucketed into sessions which denote a chunk of observations before the user switches between apps on their phone. The input data within a session is sampled every 2 seconds or with a frequency of 0.5 Hz. Since some user sessions can be arbitrarily long such as if the user is watching a video or browsing, we only take the tail of each session of maximum 60 second duration to make predictions about the user’s next action. 

In \textbf{Table \ref{tab:lerna_categories}} it can be noted that the data is a highly skewed distribution with 73\% of sessions coming from "Social” app category. This was expected as passive browsing of social media applications is prevalent amongst mobile users and thus frequent switching to the Social app category was observed. The next prevalent app category was Productivity which includes the group of apps like e-mail, calendar, docs, alarm etc.\\

\begin{table}[h]
\begin{tabular}{|c|c|c|c|c|}
\hline
\rowcolor[HTML]{EFE8A0} 
\textbf{Size} & \textbf{Categories} & \textbf{Devices} & \textbf{Features} & \textbf{No. of Days} \\ \hline
5.7 GB        & 8                   & 39               & 91                & 159                  \\ \hline
\end{tabular}
\caption{Summary of the Lerna Pilot study dataset}
\label{tab:lerna_data}
\end{table}

\begin{table}[h]
\begin{tabular}{|
>{\columncolor[HTML]{EFE8A0}}c |c|c|c|c|c|c|c|c|}
\hline
\textbf{Categories} & Audio  & Game   & Images & Maps   & News   & Productivity & Social  & Video  \\ \hline
\textbf{Frequency}       & 1,030  & 480    & 764    & 676    & 633    & 8,635        & 35,499  & 869    \\ \hline
\textbf{Proportion}      & 2.12\% & 0.99\% & 1.57\% & 1.39\% & 1.30\% & 17.77\%      & 73.06\% & 1.79\% \\ \hline
\end{tabular}
\caption{Summary of broad App Categories in User Sessions}
\label{tab:lerna_categories}
\end{table}

\section{Related Work}
Prior work in the task of Human Activity Recognition can be divided into two broad categories: hand-engineered signal processing based feature methods and Deep learning based methods. In the first category, simple models such as Logistic Regression, SVM with linear/non linear kernels, Random forest etc. are used with hand-engineered features obtained from raw acceleration, gyroscope values. The authors of the UCI dataset\cite{UCI_HAR_PAPER} for e.g. use an SVM model with an RBF kernel using features obtained by transforming time series data into frequency domain using Fourier transform. The features computed include statistics such as mean, skewness, kurtois of the frequency spectrum as well as signal energy, Interquartile range, entropy etc. They obtained accuracies of upto 90\% in their experiments.\\

For the latter class of methods, the representational capacity of models such as CNNs\cite{R2} \cite{sikder2019human} and LSTMs\cite{hernandez2019human} are leveraged for automatic learning of optimal features without manual feature engineering. In CNN models, 1-d CNNs are a favourable choice in modelling timeseries data, whereas in RNN models, Bi-directional LSTMs\cite{hernandez2019human} particularly show promising results obtaining accuracies of upto 95\% on UCI dataset. In our methodology, we detail our experiment with both feature engineering and LSTM model on the Real-life HAR dataset for this task. 

\section{Methodology}
\subsection{Part-A: Activity Recognition}

Based on prior literature, we experiment with both feature engineering and deep learning methods such as an LSTM model for the task of Activity Recognition. As concluded in the dataset section, we chose the Real-life HAR dataset consisting of 4 activity classes: Inactive, Active, Walking, and Driving, for our experiments. Since we wanted to use the predictions of the trained activity recognition model on the Lerna Pilot dataset, we pre-process the Real-life HAR dataset to have the same session duration as the data sessions recorded in the Lerna Pilot dataset i.e. \textbf{60 seconds}. Further, we sample accelerometer, gyroscope readings within a session at the same frequency as the Lerna Pilot dataset i.e. at \textbf{f=0.5 Hz}. This required writing synchronization scripts as the raw dataset contained sensor readings sampled from multiple devices with different frequencies. We use the same sliding window technique to create sessions as proposed in the paper with a 95\% overlap. The data distribution for the sessions corresponding to the 4 activity classes is plotted in \textbf{Figure \ref{fig:2.4}}. The dataset is divided into 80-10-10 for the train, val, test splits respectively. \\

 We describe our two experiment methodologies below.\\

\textbf{Experiment Summary}\\

\begin{enumerate}
    \item \textbf{LSTM Model:}  We create a simple LSTM \cite{hernandez2019human} classifier model that accepts an input of size: (Batch, 30, 6), where 30 is the sequence length based on a 60 second session sampled at 0.5 Hz = 30 observations. And, 6 is the input dimension size of the tri-axial (x,y,z) acceleration and gyroscope data respectively. The model architecture contains 2 LSTM layers with hidden dimension of 32 followed by a fully connected layer that outputs softmax probabilities for each of the 4 classes. We train the model with auto-stopping criterion for 100 epochs with Adam optimizer.\\
    \item \textbf{Feature Embeddings:} We create handcrafted features from signal processing literature consisting of both time-series and frequency components. We use the same feature set as the UCI dataset paper \cite{UCI_HAR_PAPER} described in \textbf{Table \ref{tab:features}}. Fast-fourier transform is computed over the 30 observations of the 60 second session to obtain frequency components. Both time series and frequency features are used to compute the final feature set which is of dimension 112. The obtained feature set is used as input to an SVM classifier with an RBF kernel with $\gamma = 0.1$ and $C=100$ obtained through hyperparameter search on validation set. We also compare against an SVM with a linear kernel.\\
\end{enumerate}

We detail the results obtained in our experiments in \textbf{Table \ref{tab:exp_results_har}}. We use macro F1 score as our evaluation metric which assigns equal weights to each class while averaging class wise f1 scores. \\

\newpage

\renewcommand{\arraystretch}{1.2}
\begin{table}[h]
\begin{tabular}{|l|l|}
\hline
\rowcolor[HTML]{EFE8A0} 
\textbf{Function} & \textbf{Description}              \\ \hline
mean              & Mean Value                        \\ \hline
std               & Standard Deviation                \\ \hline
mad               & Median Absolute Value             \\ \hline
max               & Largest value in array            \\ \hline
min               & Smallest value in array           \\ \hline
peaks             & No. of peak values in a session    \\ \hline
numMean           & No. of values above mean            \\ \hline
sma               & Signal Magnitude Area             \\ \hline
Energy            & Average sum of squares            \\ \hline
iqr               & Interquartile range               \\ \hline
entropy           & Signal Entropy                    \\ \hline
arCoeff           & Autoregression coefficients       \\ \hline
correlation       & Correlation coefficient           \\ \hline
minFreqInd        & Smallest frequency component       \\ \hline
maxFreqInd        & Largest frequency component       \\ \hline
meanFreq          & Frequency signal weighted average \\ \hline
skewness          & Frequency signal Skewness         \\ \hline
kurtosis          & Frequency signal Kurtosis         \\ \hline
energyBand        & Energy of a Frequency interval    \\ \hline
\end{tabular}
\caption{Description list of the computed feature set from raw acceleration, gyroscope sensor data}
\label{tab:features}
\end{table}

\renewcommand{\arraystretch}{1.2}
\begin{table}[h]
\begin{tabular}{|c|c|c|}
\hline
\rowcolor[HTML]{EFE8A0} 
\textbf{Experiment} & \textbf{Features}       & \textbf{Macro F1-score (\%)} \\ \hline
Linear SVM          & Time-series + Frequency & 57.49                        \\ \hline
SVM w/ RBF Kernel   & Time-series + Frequency & 84.79                        \\ \hline
LSTM                & Time-series             & 97.66                        \\ \hline
\end{tabular}
\caption{Results obtained for the Activity Recognition Experiments}
\label{tab:exp_results_har}
\vspace{2.5mm}
\end{table}

\begin{figure}[h]
    \centering
    \includegraphics[width=0.65\textwidth]{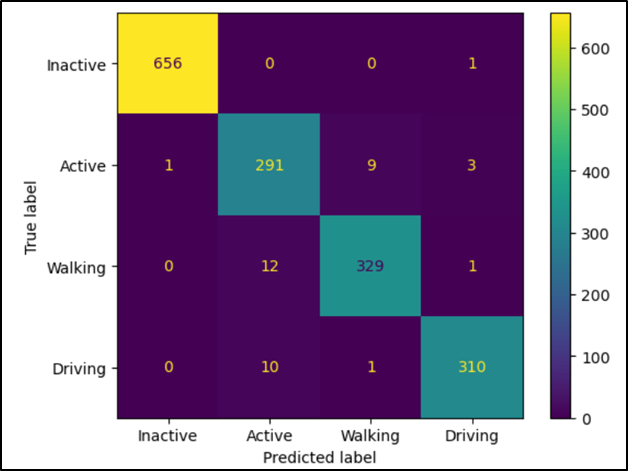}
    \caption{Confusion matrix plotted on the Test set for the LSTM model. The model is able to easily differentiate between classes obtaining a macro f1-score of 97.66 \%.}
    \label{fig:2.6}
    \vspace{-15mm}
\end{figure}

The LSTM experiment performs the best amongst the other two feature engineered SVM methods obtaining a macro-f1 score of 97.66 \%. We now use these trained models for inference on the Lerna Pilot dataset to process the raw sensor data into either a one-hot representation of the predicted class, or by converting the time-series data of 30 observations into a 112 dimensional feature embedding using our previously defined feature set in \textbf{Table \ref{tab:features}}. This is detailed in the next section below.\\

\subsection{Part-B:  Testing on a Pilot Dataset}

In this experiment, we wanted to utilize the trained LSTM model for the activity recognition task as well as the hand-crafted feature embeddings to compare their efficacy against directly inputting sensor data, to a Logistic Regression model.  This pilot study dataset was not part of the company’s main product but was a user study to understand the integration of the Lerna mobile recommendation library. The task was a dummy objective which predicted what application the user is going to open next on their device based on the sensor data and the device context. The baseline experiment model for this pilot dataset was the Logistic Regression model, which was already implemented for mobile deployment as part of the recommendation library. \\

Our hypothesis for this experiment was that for simple linear models such as Logistic Regression, it would be better to process raw accelerometer and gyroscope sensor data into predicted activity labels, or compressing the raw time-series session into feature embeddings. The complete input set for this experiment corresponds to the Lerna SDK Sensor suite data described in \textbf{Table \ref{tab:sensors}}. The sensor suite contains both numerical, categorical and multi-valued features. The numerical features corresponding to the accelerometer and gyroscope data are processed into Activity data either as predicted labels or feature embeddings, while the other categorical data, multi-hot feature set is encoded as one-hot/multi-hot binary vectors respectively. \\

It is to be noted, that in this experiment we were not interested in finding the best model for this dummy objective as that was already concluded in previous year experiments, but rather answering the question of what’s a better input representation of the activity sensor data with our baseline model Logistic Regression which also serves as the baseline for our federated mobile recommendation library. With that in mind, we list our experiments below.\\

\textbf{Experiment Summary}
\newline

\begin{enumerate}
    \item \textbf{Individual data point prediction:} Since Logistic Regression model doesn’t input sequential data observed in a user session of 60 seconds, we predict the next app category class at each data point independently and then take mode of the predicted class. Thus, the model at each instance only observes the current sensor state without any notion of history. The model makes prediction at the same frequency as input, 0.5 Hz or every 2 seconds and the mode of 30 predictions is taken as the prediction for the session. \\
    
    \item \textbf{Session Activity label:} In this experiment, we use our trained LSTM model from PART-A of our experiments to predict a single activity label for a 60 second user session. The activity label along with mean/mode statistics of other numerical/categorical device data respectively,  which vary infrequently during a session are used as input as one single vector to a Logistic Regression model. The output class is the predicted category for the entire session. \\
    
    \item \textbf{Session Activity Embedding:} In this experiment, we compute a 112 dimensional feature embedding for the activity sensors based on \textbf{Table \ref{tab:features}}. Similar to the LSTM experiment, a single prediction is made for the session based on the input vector consisting of activity embeddings, mean/mode statistics of other device data, to the Logistic Regression model. \\
\end{enumerate}

Since the dataset is imbalanced towards the “Social” category class, we use the Balanced accuracy metric \cite{scikit_learn_model_evaluation} from scikit-learn library which averages class accuracies weighted by their inverse prevalence. The same metric was used in previous year experiments and thus allows us to compare the baseline experiment: Individual data point with LR against our two new experiments.  We also use the Dummy classifier as a heuristic baseline which simply predicts the majority class every time ignoring the input features.\\

\begin{table}[h]
\begin{tabular}{|c|c|}
\hline
\rowcolor[HTML]{EFE8A0} 
\textbf{Experiment}                 & \textbf{Balanced Accuracy(\%)} \\ \hline
Dummy classifier                    & 12.5 \%                        \\ \hline
Individual Data point               & 13.25 \%                       \\ \hline
Session Activity label              & 12.95 \%                       \\ \hline
\textbf{Session Activity Embedding} & \textbf{15.13 \%}              \\ \hline
\end{tabular}
\caption{Results comparing different Experiments on the Lerna Pilot dataset}
\label{tab:exp_results_partB}
\end{table}

To our surprise, we observe that passing the activity label as an input directly using the LSTM prediction performs objectively worse than the individual data point experiment. Computing a condensed feature embedding representing the activity sensor data however, beats previous baseline and performs the best. It can be observed however, that the task of predicting the user intent based on the mobile device and sensor data itself, is objectively hard and requires further context. To give a fair comparison with previous year experiments, the best performing model on this task previously was an XGBoost model with a Balanced accuracy of 19.60. We discuss our conclusion for both parts of our experiments in the section below.\\

\section{Conclusion}

In this chapter, we discussed our methodology towards harnessing the sensor and device information data from user’s mobile devices. We listed the sensor sources which we term as mobile context data involved in our downstream recommendation pipeline. We conducted experiments to process the raw data from accelerometer, gyroscope for the task of Human Activity Recognition. In absence of private company datasets, we used public datasets such as Real-life HAR dataset to train our models and use them for inference on downstream tasks. We tested these methods for a dummy task of user intent prediction based on the dataset from a previous pilot study conducted by the company. Overall, we took home the following valuable learnings for implementing the next parts of our pipeline and made some key contributions on the engineering end for deployment of the experimented models.  \\

\begin{itemize}
    \item Based on the results obtained in our pilot study experiment, we concluded that processing raw activity sensor data into feature embeddings greatly helps in encoding the mobile context for the task. The activity labels encoded as binary one-hot labels on the other hand harm the Logistic Regression model performance, which we hypothesize could be because LR model is suited to learn feature combinations between different sensors better on continuous data, rather than boolean.\\
    \item Previously our LR model was making individual prediction every 2 seconds, whereas under our new methodology we experimented with session level predictions by encoding such sequential data into one single embedding. This had a direct ramification in the company’s deployed model and became our de-facto methodology. For our downstream recommendation pipeline, we now use the same processing methodology for our input mobile sensor data.\\
    \item To our surprise, we didn’t find library implementations of Fast Fourier transform in Kotlin Multiplatform, for computing these activity feature embeddings in our mobile SDK on Android/iOS devices. Therefore, a good chunk of time was devoted towards writing efficient implementation of FFT and all other statistics such as signal energy, correlation etc. from scratch in Kotlin as part of the engineering efforts.\\
\end{itemize}

\chapter{Two Stage Recommendation System}

\section{Introduction}
In this chapter, we explain our approach towards building a hyper-personalized recommender system for mobile devices. We divide our system into two-stages, a centralized stage that is deployed on cloud servers and a federated stage which runs on user’s mobile devices. In the following sub-sections, we categorize the various data sources involved in the system and briefly explain each component of the two-stage recommendation system and our motivation for splitting the pipeline into a central and a federated stage. \\

Further in this chapter, we do a literature survey of existing recommendation system methods, the datasets used, our methodology and experiments for the first or Centralized stage of our pipeline. The second stage of our recommendation system which is the federated stage is explained in the next chapter of this work.

\subsection{Data categories}
\label{subsection: 3.1.1}

We categorize the data involved in building our hyper-personalized recommendation system into three groups which are detailed as following.\\

\begin{enumerate}
    \item \textbf{App-context data:} This data is collected by the app publisher about the user and their interactions within their platform. We further split this data into two sub-groups as: \\
    \begin{itemize}
        \item \textbf{User data}: This consists of demographic data that the user manually enters in the app signup process such as their name, age, gender, location etc. And further,  preferences data such as music or movie genres which for example are collected by apps such as Netflix or Spotify as part of their signup process to handle the cold start problem.
        \item \textbf{Interaction data}: This consists of data about user-item interactions such as purchase history, click or view history, Wishlist or cart items etc.\\
    
    \end{itemize}
    We group all such data collected by the app publisher on their platform(web or mobile) itself as the app-context data.\\
    
    \item \textbf{Mobile-context data}: The data that originates solely from the mobile device such as sensor data, device state, user habits data etc. under the context of the app usage is called mobile-context data. This data is collected as part of the Lerna SDK and not the app-publisher itself. A detailed breakdown of all the input sources of this data is already discussed in \textbf{Section \ref{sec:2.2} Table \ref{tab:sensors}}. This data is located on the mobile device and under our federated learning paradigm, never leaves the user’s device.\\

    \item \textbf{Item data}: This consists of the metadata of items or products being recommended to the users. For eg, a movie recommendation app would have item metadata consisting of genre, release year, title, summary, ratings etc.\\
\end{enumerate}

\subsection{Two-stage Recommendation pipeline}
In this work, we propose a two-stage system that can be utilized by any mobile app developer to offer hyper-personalized recommendations to their users. The proposed solution is split into two stages. The first stage is a centralized collaborative filtering recommendation algorithm deployed on cloud servers, and the second stage is a federated Click-through-rate Prediction algorithm deployed on user mobile devices, which is used for ranking items suggested in the first stage by hyper-personalizing them to user activity.\\

\begin{figure}[h]
    \centering
    \includegraphics[width=0.97\textwidth]{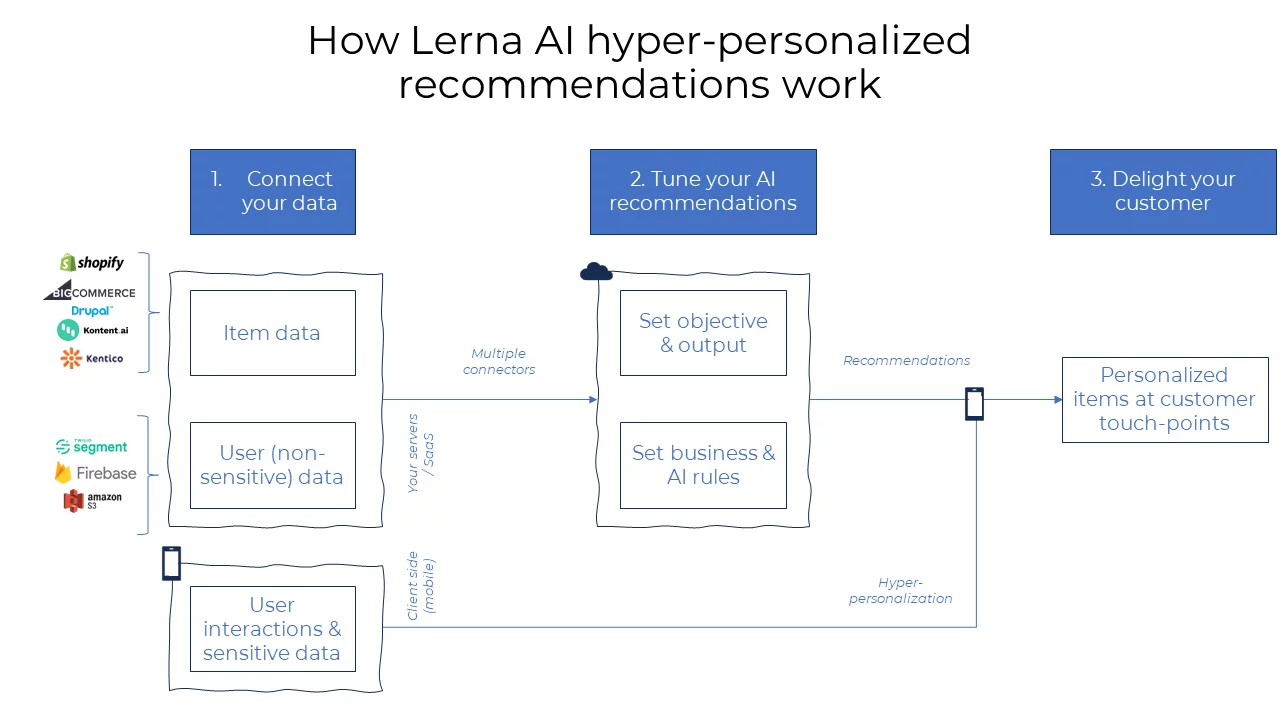}
    \caption{The proposed two-stage Recommendation pipeline}
    \label{fig:3.1}
\end{figure}

Our pipeline is described in \textbf{Figure \ref{fig:3.1}} detailing the input data source integrations for each stage.\\

The first stage, known as the centralized stage, takes item metadata along with non-sensitive user data such as user preferences and item interactions, collectively referred to as App-context data, as described in the previous section. We describe the Correlated Cross-Occurrence based collaborative filtering algorithm deployed in our system in detail in the next sections of this chapter. \\

In the second stage, which is federated, sensitive user activity data, referred to as Mobile-context data, obtained from their devices, is used along with some item data to rank the recommended items from the previous stage and show the Top-k relevant items to the user. The algorithm also uses the recommendation score predicted in the first stage as an input feature for the next stage. 

The algorithm used is an Attention-transformer based machine learning model that assigns a score between 0 and 1 to every item, which is then used to get a Top-k ranking of items. More details about the second stage will be discussed in the next chapter of this report in \textbf{Chapter \ref{chap:4}}.\\

\textbf{Motivation for a two-stage system:}
\newline

The main motivation for implementing a two-stage pipeline for our system is to give the company’s end customers i.e. the App developers flexibility to plug-in their existing recommendation system in the first-stage. They can then further improve their recommendations by opting for personalization to user’s mobile habits using only the second federated stage of our system.

This two-stage recommendation system can therefore be used either as a standalone solution or as an extension to an existing system. This approach reduces the friction involved in onboarding new apps, which is why we chose this pipeline instead of directly using a single federated recommendation system, for example the federated collaborative filtering algorithm \cite{ivannikova2019federated}.\\

\section{Literature Survey}

Large-scale Recommendation systems which address millions of users and product offerings are generally built in stages, where each stage progressively filters the product or content that is relevant to a given user. These stages are defined as following:\\

\begin{enumerate}
    \item \textbf{Candidate generation:}  In this first stage, the system starts from a potentially huge corpus and generates a much smaller subset of candidates which could be relevant for the user. Traditional recommender system methods such as collaborative filtering, content-based method and hybrid methods fall under this category. 
    \item \textbf{Scoring \& Ranking:} In this stage, candidate items are scored and ranked based on user attributes and business rules. Click-through-rate prediction models which generate a probability score for the likelihood of an item being bought or viewed fall are generally used in this stage. Ranking models which compute pairwise or group relevance of items also fall under this category.
    \item \textbf{Re-ranking:} This final stage of recommendation system is essential only for sophisticated large scale systems which require adjustments for fairness, diversity \& freshness in the top results. In this stage, the ranked items from previous stage are re-ranked based on the above criterion.
\end{enumerate}

Drawing parallels with our system, the first stage of Candidate generation corresponds to the Centralized recommendation system stage of our system, whereas the Scoring \& Ranking stage corresponds to the second stage of our system, which is a federated Click-through Rate (CTR) prediction model. We do a literature survey of the Candidate generation methods in this chapter and of federated learning and CTR models in the next chapter.

\subsection{Candidate Generation Models}

Candidate generation Recommendation systems are commonly categorized into three main types: Collaborative Filtering based, Content based, and Hybrid. We discuss some popular methods for each below:\\

\begin{enumerate}
    \item \textbf{Collaborative Filtering based:} The basic idea in these methods is to predict user’s preferences by leveraging the collective knowledge of a large pool of users. User’s interactions such as ratings or buy history are recorded in a user-item interaction matrix which is used to suggest new items to users.  These methods are further classified into two categories: memory-based and model-based.\\
    \begin{itemize}
        \item \textbf{Memory-based methods:} These methods compute similarity metrics between users or items based on the interaction matrix and recommends items based on sorted similarity values. The method involves computing nearest neighbours based on a similarity threshold to offer user-based or item-based recommendations. Metrics such as Cosine similarity, Pearson Correlation Coefficient, Co-Occurence probability etc. are used for computing the similarity. These methods are simple to implement and perform well in encoding historical user preferences but encounter challenges in large state spaces. Sparsity of the interaction matrix is handled by computing embeddings that transform high dimensional sparse interaction vector to a low dimensional dense space. Method such as prefs2Vec \cite{valcarce2019collaborative} inspired by the Word2vec NLP model compute user/item embeddings before computing similarity. Graph based methods such as \cite{chen2019collaborative} are also used to compute embeddings.\\
        \item \textbf{Model-based methods:} These methods use parametrized models such as neural networks to model latent relationships between users and item preferences. Matrix factorization methods are the most popular class of methods in this category. Neural Collaborative Filtering \cite{he2017neural}, Neural matrix factorization \cite{dziugaite2015neural} and Factorization machine methods such as DeepFM \cite{guo2017deepfm} are some example of popular methods in this category. \\
    \end{itemize}

    \item \textbf{Content-based:} These methods rely on item metadata information such as text descriptions, categories, genres etc. to solve user-specific classification problems on item features to provide personalized recommendations. These methods do not rely on other user’s ratings to recommend items and therefore do not suffer from a cold-start problem unlike collaborative filtering methods. Multiple modalities such as text, audio, image features can be used to build item features based on the task. Some recent works also included leveraging knowledge graph to build features such as Yang et al. \cite{yang2022personalized} and  Deldjoo et al. \cite{deldjoo2018using} which leveraged visual features for movie recommendations.\\

    \item \textbf{Hybrid methods:} These methods leverage both content-based and collaborative filtering methods either as ensemble or in conjunction in stages to provide domain specific accurate recommendations. Polignano et al. \cite{polignano2021together} for e.g. blends user-item interactions and contextual graph features for social recommendations.\\
\end{enumerate}

\section{Centralized Recommendation System}
\label{sec:3.3}

When we investigated methods for building the first stage of our pipeline, there were the following key considerations:

\begin{itemize}
    \item The implemented centralized stage must be a general purpose recommendation system that works with multiple app domains such as news articles, games, music recommendations etc. Therefore the company’s focus was to prioritize ease of integration and generalizability for our end customers which are app-developers instead of building a highly performant recommendation system which is tailored for a specific use case.
    \item Since the proposition of the company Lerna AI is to provide recommendations as an API service to app-developers, the proposed system should be highly scalable and support integrations with popular large scale databases such as MongoDB, MySQL, Elasticsearch etc.
    \item The system should support diverse inputs such as user profile data, user-item interactions, item metadata etc without making strong assumptions on their data type.
    \item The recommendation system should have a low throughput and should support adding hard-coded business rules.\\
\end{itemize}

\subsection{Universal Recommender}

Under these considerations, the best recommendation method for our use-case was a memory-based collaborative filtering algorithm called the Universal Recommender \cite{universal-recommender} developed by the ActionML \cite{actionml} team. The algorithm proposed in this system is a Correlated Cross-Occurence algorithm which computes log-likelihood statistic as a similarity metric to offer multiple styles of recommendations. These include item-to-item (given this item recommend similar items), user-to-item (given user history, recommend an item) and item-set recommendations (given a set of items recommend other items, used in shopping cart, wish lists etc.) Further, heuristics based recommendations such as popularity or top items, business logic based filters are also supported in this system. \\

The reason why the authors brand it as a universal recommender system is because it’s able to ingest any number of user actions, events, profile data, and contextual information. It then serves results in a fast and scalable way using the Apache Spark \cite{spark} backend. While the system doesn’t process item metadata such as text to compute embeddings the way content based recommendation systems do, it does support setting text based filters on metadata and adding business rules for boosting recommendations based on item properties. In this regard, then system can be viewed as a hybrid recommendation system utilizing both user interactions and item properties. 
\newline

\textbf{Primary and Secondary Indicators:}
\newline

Traditional matrix factorization-based methods, such as Alternating Least Squares \cite{hu2008collaborative}, rely on a primary interaction event between the user and the item to construct the interaction matrix. This interaction can either be explicit feedback, such as a rating or a purchase event, or it can be implicit feedback, such as viewed items or search terms. The resulting interaction matrix is then factorized into latent user and item vectors, which represent user preferences and item properties implicitly. \\

\begin{figure}
    \centering
    \includegraphics[scale=0.6]{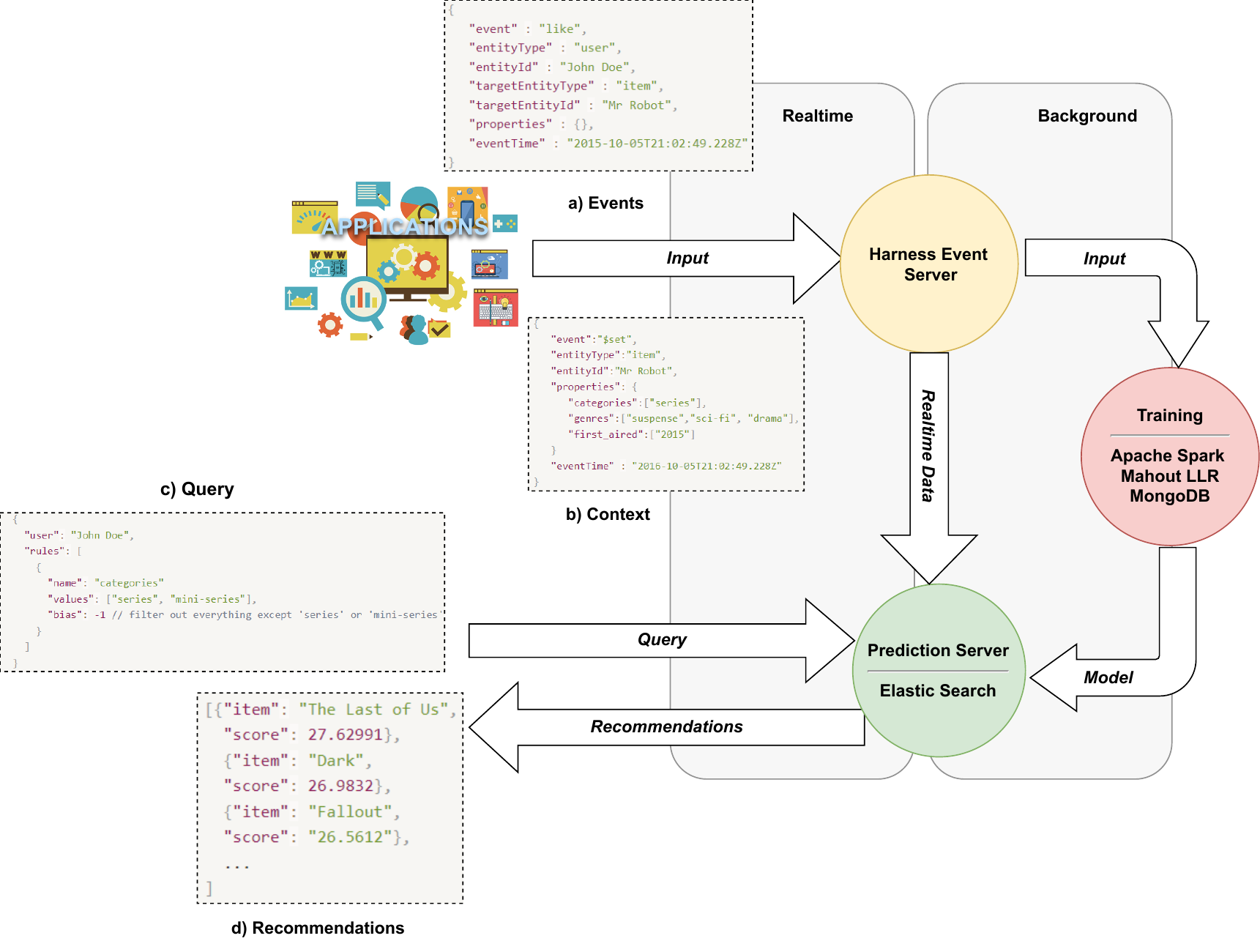}
    \caption{System Architecture diagram detailing various components of the Universal Recommendation System. The input to the system is comprised of a)Events json containing user-item interactions such as like/purchase etc. and b) Context json containing user/item properties. A sample query to the system is defined in c) Query json and the output d) Recommendations json are served along with their computed Log-likelihood similarity scores}
    \label{fig:ur diagram}
\end{figure}

However, a limitation of this approach is that typically a single primary event is used to construct the interaction matrix. In the case of explicit feedback such as user ratings, this can lead to sparsity issues. Therefore, methods such as the Universal Recommender propose to tackle this issue by incorporating both explicit(primary) and implicit feedback(secondary) events into a single cohesive formulation. This flexibility to ingest any number of user interactions whether primary or secondary was the motivation to utilize this method in our system. A system architecture diagram detailing all components of the Universal Recommender is drawn in \textbf{Fig-\ref{fig:ur diagram}}. 

We further describe the co-occurence algorithm used in this method in the next section below.

\subsection{Correlated Cross-Occurrence algorithm}

The algorithm used in the Universal Recommender \cite{universal-recommender} is a similarity based recommendation method where items are scored based on the co-occurrence between events. This means that while there is an interaction matrix involved, there is no matrix factorization done to compute latent user-item embeddings. Let us understand the different terminologies involved in the algorithm:\\

\begin{enumerate}
    \item \textbf{Cross-Occurence:} In recommender systems we often need to recommend items based on some notion of similarity, e.g. people who viewed/bought item A, also viewed/bought item B. If we consider the case where we are interested in “user bought A, would they buy B?” then this would be an example of computing the co-occurence where the event remains same P(bought B | bought A) i.e. both are primary events.
    
    On the other hand, if we consider the case of “viewed A, bought B?”, then this would be an example of Cross-Occurrence where we are making recommendation on primary event “bought” based on secondary event “view”.  The key behind the Universal Recommender system is to incorporate all user actions to compute these Cross-occurence scores against the primary indicator.\\

    The actual computation of the naive Cross-occurence metric is fairly simple and can be defined as: \\
    
    $\begin{aligned}
    p(\text { bought } B \mid \text { viewed } A) & =\frac{\text { number of people who bought } B \text { and viewed } A}{\text { number of people who viewed } A} \\
    & =\frac{\text { cooccurrence }}{\text { number of people who viewed } A} 
    \end{aligned}$ \\ \\

    However co-occurrence based similarity suffers from the problem of repeatedly recommending items that are globally popular since they will naturally have the highest co-occurences with most items. Instead, what we want are recommendations that are specific to each item, also called anomalously relevant items. This is where the log-likelihood statistic comes into use which we discuss next.\\

    \item \textbf{Log-Likelihood Ratio:} The “correlated” part of this Correlated Cross-Occurence algorithm comes from computing the log-likelihood ratio which is nothing but the statistical Likelihood-ratio test \cite{note1982likelihood}.  The log-likelihood ratio is a similarity score that does not only depend on the number of times two events have occurred together but also on the number of times two events haven’t occurred together and the number of times only one event has occurred and the other has not. The score will be higher if there is a correlation or anti-correlation between two events.

    It is computed by defining the conditional probability distributions of these events. The key idea being that we wish to find anomalously relevant items by first defining a null hypothesis which states that “is viewing item A and purchasing item B independent processes?”,  i.e. $P(bought B|viewed A) \approx P(bought B|not viewed A) \approx P(bought B)$. \\

    If that is the case, items A and B are unrelated, and should not be recommended together. Conversely, if these probabilities are very different, this indicates an anomalously relevant pair of items which should be recommended together. A detailed discussion of computing the log-likelihood ratio is out of scope of this chapter, however Adobe’s Data science team has an excellent tutorial \cite{adobe_llr_paper} explaining use of log-likelihood ratios for co-occurences which this is referenced from.\\
\end{enumerate}

\textbf{Algorithm Definitions}
\newline

We borrow the algorithm description of the Correlated Cross-Occurence algorithm from the ActionML author's presentation slides(\cite{ferrel_unified_recommender} explaining the Universal Recommender system. We reference it as below.  \\

Let $r=$ recommendations,

$h_p=a$ user's history of some primary action (purchased item or liked movie for instance) 

$P=$ the history of all users' primary action rows are users, columns are items

$[\mathrm{P^{T}} \mathrm{P}]=$ Column-wise item similarity using  using the log-likelihood based co-occurrence score.

Virtually all existing collaborative filtering type recommenders use only one indicator of preference-one action
$$
r=h_p\left[P^{T} P\right]
$$

The key proposal however, of the Correlated Cross-occurence algorithm is to ingest all user actions containing secondary indicators, for e.g. Viewed items, Category preferences etc. by computing Cross-occurences scores with the primary indicator. This is denoted by: $\left[V^{T}P\right]$ and $\left[C^{T}P\right]$ respectively in the below equation.\\

Further, user and item properties are also encoded in the same formulation by defining any custom similarity metric such as cosine similarity to compute similarity between content tags for example. This is denoted by: $\left[TT^{T}\right]$, where T denotes an item content tag.\\    

The final  equation therefore becomes:
$$
r=h_p\left[P^{T} P\right]+h_v\left[V^{T} P\right]+h_c\left[C^{T} P\right]+ h_t
\left[TT^{T}\right] + \ldots
$$ 

where $h_t$ similarly denotes a user's history of an action on items with
tags.\\

Universal Recommender under the hood uses Apache Mahout \cite{mahout} which is a distributed and highly scalable machine learning library to compute the likelihood-ratio co-occurences in a highly parallelized and efficient manner. The library implements certain improvements such as the MapReduce algorithm proposed in \cite{schelter2012scalable} for computing the pairwise item co-occurences which scales with linearly with number of users making it efficient for large scale recommendation tasks.\\

\section{Datasets}
\label{sec:4.4}

For our task of building a mobile-first recommendation system, we were looking for public datasets that fit the categorization specified in \textbf{Section \ref{subsection: 3.1.1}}  which we would observe in our real world data stream. This includes user features such as user profile data, user-item interactions, item features and most importantly mobile specific features such as device and sensor activity data. \\

However, this proved to be highly challenging and we didn’t find any one single dataset that directly fit the profile of our use-case. The available datasets that included mobile sensor data \cite{endomondo_dataset}\cite{UCI_HAR} contained only user features and no item level interactions or metadata. Therefore, in absence of company’s own private data, we simplified our problem and focussed on datasets that broadly represent our categories such as having user-level features and item-level features. Sensor data  obtained from mobile devices can be assumed to be another dimension to user level features. Based on this understanding, we explored options to evaluate the Universal Recommender system detailed in \textbf{section \ref{sec:3.3}} and our following Federated learning stage using public datasets containing these two sets of features.\\

We explored multiple datasets such as the Yelp ratings dataset \cite{yelp_dataset}, Taobao Recommendation dataset \cite{Taobao_Recommendation}, RetailRocket e-commerce dataset \cite{roman_zykov_noskov_artem_anokhin_alexander_2022} and the famous MovieLens 1M dataset \cite{movielens_dataset}. All of these datasets contained both user features and item features. Since we wanted to evaluate both stages, the collaborative filtering based Universal Recommender system and our federated Click-through-rate prediction based algorithm, we looked for prior literature containing some baselines with these datasets for us to compare against. We found that MovieLens 1M dataset aptly fit our use case with prior literature and baselines to compare our experiments for both stages. Therefore, we used this dataset for our experiments in both centralized and federated stage.\\

\subsection{MovieLens 1M dataset}
The MovieLens 1M dataset was released by the GroupLens research group at the University of Minnesota in February 2003. It consists of total 1M movie ratings in the integer range of 1 to 5. The total exact number of users in the dataset are 6,040 with 3,706 rated movies. The dataset contains total 7 input feature fields consisting of both user level features and item level features explained below. \\

A brief description of the dataset fields is given below:
\begin{enumerate}
    \item "movie\_id": a unique identifier of the rated movie
    \item "movie\_title": the title of the rated movie with the release year in parentheses
    \item "movie\_genres": a sequence of text genres to which the rated movie belongs.
    \item "user\_id": a unique identifier of the user who made the rating.
    \item "user\_rating": the score of the rating on a five-star scale.
    \item "timestamp": the timestamp of the ratings, represented in seconds since midnight Coordinated Universal Time (UTC) of January 1, 1970
    \item "user\_gender": gender of the user who made the rating; a true value corresponds to male
    \item "bucketed\_user\_age": Bucketed age values of the user between 18-56+ age range. 
    \item "user\_occupation": the occupation of the user who made the rating represented by an integer-encoded label
    \item "user\_zip\_code": the zip code of the user who made the rating

\end{enumerate}

\section{Evaluation Metrics}

Below are the two metrics that we have used to evaluate the performance of Universal recommender system:

\begin{enumerate}
    \item \textbf{HitRate@k:} It is 1, if at-least one relevant item is recommended within the top k items, otherwise it's 0. We compute the mean across all users as the final hit ratio score. This is a key metric for us in evaluating the performance of our candidate generation recommendation model. 
    It is mathematically defined as follows:

$$HR@k=\frac{\left|U_{h i t}^k\right|}{\left|U_{a l l}\right|}$$

    where $\left|U_{h i t}^k\right|$ is the number of users for which the correct answer is included in the top k recommendation list, $\left|U_{a l l}\right|$ is the total number of users in the test dataset.
    
    \item \textbf{NDCG@k:} It stands for Normalized Discounted Cumulative Gain and is a metric for comparing the ranking quality of the predicted items. It compares rankings to an ideal order where all relevant items are at the top of the list. NDCG\@K is determined by dividing the Discounted Cumulative Gain (DCG) by the ideal DCG representing a perfect ranking. It is a standard metric in evaluation of recommendation and ranking systems. \\
    
    \begin{align*} NDCG@K = \frac{DCG@K}{IDCG@K} = \frac{\sum_{i=1}^{k} \frac{gain_i}{\log_2(i+1)}}{\sum_{i=1}^{k} \frac{gain_i^{ideal}}{\log_2(i+1)}} \end{align*}
    
    However, for our first stage of candidate generation model we are less concerned about the order of ranking for the predicted items and thus this is a secondary metric for the evaluation of the Universal Recommender system. 
    
\end{enumerate}

\section{Experiments}

We conducted experiments to evaluate the performance of the Universal Recommender system on MovieLens 1M dataset. We couldn’t find any publication or publicly available benchmark by the authors of the ActionML team comparing the method’s efficacy on this dataset. Therefore, we evaluate the method ourselves on a Top-k recommendations task with metrics such as Hit Rate@k, NDCG@k detailed in the evaluation metrics section above.\\

\vspace{5mm}

\textbf{Pre-processing Inputs for Recommendation Engine:}
\newline

The ingestion format of the Universal Recommendation engine require that we convert user-item interactions such as ratings into primary and secondary indicators. Primary indicator is a singular type of interaction event that is a very clear indication of user preference (such as products bought, or movies liked) and secondary indicators correspond to all other forms of interactions that we think might add context (viewed products, user preferences such as dislikes, search terms, us etc.) The Universal Recommender is built on a distributed Correlated Cross-Occurrence (CCO) Engine, which basically means that it will test every secondary indicator to make sure that it actually correlates to the primary one and those that do not correlate will have little or no effect on recommendations.\\

We therefore convert MoviLens 1M dataset ratings ranging from [1-5] into these primary and secondary indicator events such as 
the primary indicator consists of only liked events (ratings of 
$>=4$) and secondary indicators consists of all other information in the dataset such as disliked movies (ratings $<3$), neutral (rating $=3$), user properties(gender, age, occupation), item properties(movie year, genre, title etc). We conduct experiments to understand the effect of introducing each category of these secondary indicators (events, user properties, item properties) in the output Top-k recommendations. \\

\textbf{Evaluation Methodology:}
\newline

We follow the same processing methodology used by the benchmark paper: Neural Collaborative Filtering (NCF) \cite{he2017neural} of leave-one-out evaluation where each user’s latest interaction is held out as test set and the rest are used for training. We use the same version of MovieLens 1M dataset where each user has at least 20 ratings and the rest are filtered out. We use the Github repository \cite{hexiangnan_ncf} by the authors to obtain their results for benchmark comparison. We evaluate our experiments based on the metrics of Hit Rate@Top-k and NDCG@Top-k, where k=10 items. \\

We conduct the following experiments to evaluate the Universal Recommendation Engine:\\

\begin{enumerate}
    \item \textbf{Popular Recommendations} (PopRec): This is a purely heuristics  baseline experiment where items are ranked by their popularity judged by the number of liked events (primary indicator). This is a non-personalized method and we used the implementation of Item popularity available in the Universal Recommendation(UR) Engine. \\
    \item \textbf{CCO w/ Events data}: In this experiment, we test the efficacy of Correlated Cross-Occurence which is the main recommendation algorithm of the UR Engine, using only the ratings data as primary(liked movies) and secondary indicators(disliked, neutral rated movies).\\
    \item \textbf{CCO w/ Events + Item properties:} In this experiment, we test efficacy of CCO algorithm using both events data and item properties. Similar to before, primary indicator is the liked movie event and secondary indicators are: disliked/neutral rating events, movie title, genre and release year.\\
    \item \textbf{CCO w/ Events + Item + User properties} In this experiment, we add all sources of data in the secondary indicator events to provide context for recommendation. Like before, the primary indicator remains same, whereas the secondary indicators now form the set of:  disliked/neutral rating events, movie title, movie genre, release year, user gender, user age, user occupation and their zip code.\\
    \item \textbf{Comparison Baselines:} We compare the performance of the UR Engine against methods from prior literature cited in \cite{he2017neural}. In particular, we compare against ItemKNN \cite{sarwar2001item} which is a similarity based collaborative filtering method, Bayesian Personalized Ranking BPR \cite{rendle2012bpr} which is Matrix factorization method with a pairwise ranking loss and further with the Neural Collaborative Filtering paper \cite{he2017neural} itself. We used the authors’ GitHub repository \cite{hexiangnan_ncf} to re-run these experiments and report its results below.\\
\end{enumerate}

\begin{table}[]
\begin{tabular}{|c|c|c|c|}
\hline
\rowcolor[HTML]{EFE8A0}
\textbf{Method} & \textbf{Input}                            & \textbf{Hit Rate@10} & \textbf{NDCG@10} \\ \hline
PopRec          & Events                                    & 0.380                & 0.1822           \\ \hline
CCO             & Events                                    & 0.572                & 0.2073           \\ \hline
CCO             & Events + Item Properties                  & 0.609                & 0.2159           \\ \hline
\textbf{CCO}    & \textbf{Events + (Item, User) Properties} & \textbf{0.614}       & \textbf{0.2193}  \\ \hline
ItemKNN         & Interaction Matrix                        & 0.552                & 0.3470           \\ \hline
BPR             & Interaction Matrix                        & 0.680                & 0.4200           \\ \hline
NCF             & Interaction Matrix                        & 0.730                & 0.4470           \\ \hline
\end{tabular}
\caption{Experiments comparing the efficacy of UR Engine with different inputs and benchmark methods from prior literature. }
\label{tab:experiments}
\end{table}

We discuss about the results obtained and the conclusion in the next section below. 

\section{Conclusion}
Our expectation from the Universal Recommendation Engine was to have a well-architected software system that ties up all parts involved in serving real-time recommendations. The moving parts involved in this system involve a scalable REST API web server, Database connectors to MongoDB to dump input events, Elastic Search query engine for text based filtering/retrieval and finally the Apache Mahout based Correlated Cross-Occurence algorithm. \\

Our motivation therefore, was to have a centralized Candidate generation system that accommodates all these aspects with the least amount of friction. The goal was not to have a state-of-the-art recommender system but rather a simple, reliable system that performs sufficiently well for the next stage of our pipeline. Through our experiments, we concluded that the UR Engine addresses this requirement. We discuss the results obtained below.  \\

As evident from \textbf{Table \ref{tab:experiments}}, CCO algorithm sits 
comfortably between simple heuristics based models such as
Popularity based Recommendations and complex collaborative
filtering methods such as Neural Collaborative Filtering. We
observe that augmenting the CCO algorithm with secondary inputs
such as item properties \& user profile data helps to refine
the recommendations by understanding the correlations in the data.
The performance jump was higher after inclusion of item
properties to events in comparison to inclusion of user
properties to the input. This makes sense as user data such as
age, occupation, zip code would have little correlation with
user’s taste in movies as compared to fields such as movie
genre.\\ 

The best performing CCO experiment indicates that at least one movie that user would highly like(rated 4+), was present in the top-10 items for 61.4\% of users. (Hit Rate@10 = 0.614). The Hit Rate metric is more relevant for us as compared to the NDCG metric which penalizes the relative position of the item in the list.  The best performing model from literature experiments was NCF with a Hit Rate@10 = 0.73. Overall, based on the experiment results the UR Engine is a good fit for the centralized stage of our system.

\chapter{Federated Ranking System}
\label{chap:4}

\section{Introduction}

In this chapter, we detail our proposed approach for the second or federated stage of our pipeline which involves sensitive user data from mobile devices. This step is the hyper-personalization step of our pipeline where candidate items recommended by the first stage are ranked based on the mobile context data(defined in \textbf{Section \ref{subsection: 3.1.1}} obtained from user device and the item metadata. The company’s previous method was a federated Logistic Regression model which inputs only mobile context data to provide a ranking score between [0,1] for a list of items.  We improve upon the earlier approach with a Self-attention based CTR prediction model called AutoInt proposed in \cite{song2019autoint}, which is capable of learning higher order interactions between the inputs to produce better ranking results. 

In the chapter, we discuss the following items:\\

\begin{enumerate}
    \item Federated Averaging algorithm or FedAvg \cite{mcmahan2017communication} that is used to aggregate the local CTR models trained on user’s mobile devices.\\
    \item We conduct a brief literature survey of CTR models and discuss the architecture and methodology of AutoInt\cite{song2019autoint} CTR model.\\
    \item We then conduct experiments using the MovieLens 1M \cite{movielens_dataset} dataset comparing the performance of AutoInt model against a Logistic Regression model which was the company’s previous baseline. We first conduct a centralized evaluation of these models where there is no federated aggregation. And then conduct federated learning experiments using the Flower framework \cite{flower_ai} by splitting the dataset into multiple splits, where each split represents the local data of the client device.
\end{enumerate}

\section{Federated Averaging}

In the paradigm of Federated Machine learning, data is spread across multiple devices and model training is enabled by aggregating local model parameters across devices. Models are trained locally on edge devices ensuring data privacy and security. Multiple strategies have been proposed to aggregate the model parameters to form a global model in this setting. A foundational strategy for aggregating local models was proposed by McMahon et al. \cite{mcmahan2017communication} in 2017 called Federated Averaging, or FedAvg. We use this algorithm to implement our Federated Learning pipeline and describe it briefly below.

\begin{figure}[h]
  \centering
  \begin{minipage}[b]{0.5\linewidth}
    \includegraphics[width=\linewidth]{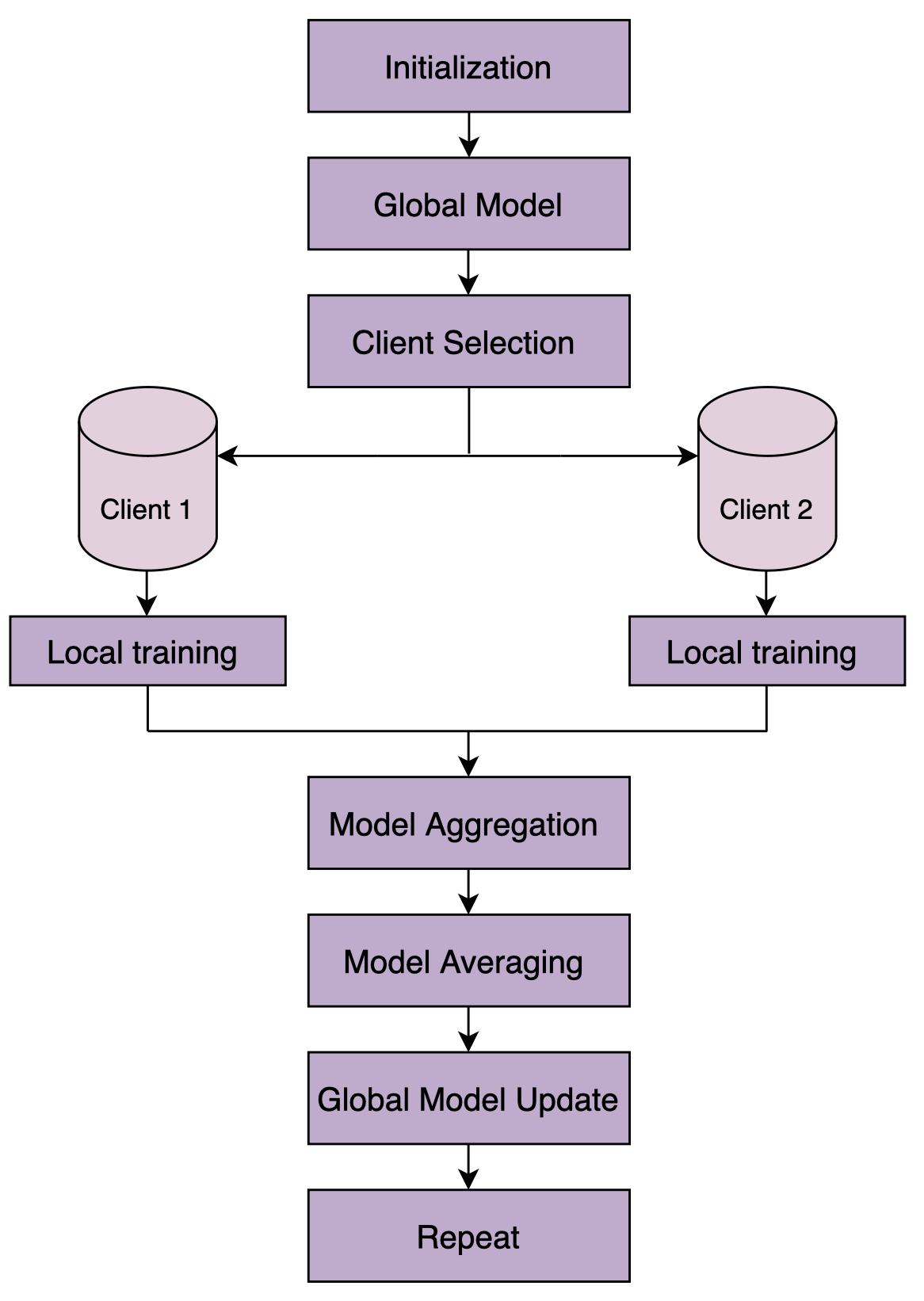}
    \label{fig:sub1}
  \end{minipage}
  \hfill
  \begin{minipage}[b]{0.48\linewidth}
    \includegraphics[width=\linewidth]{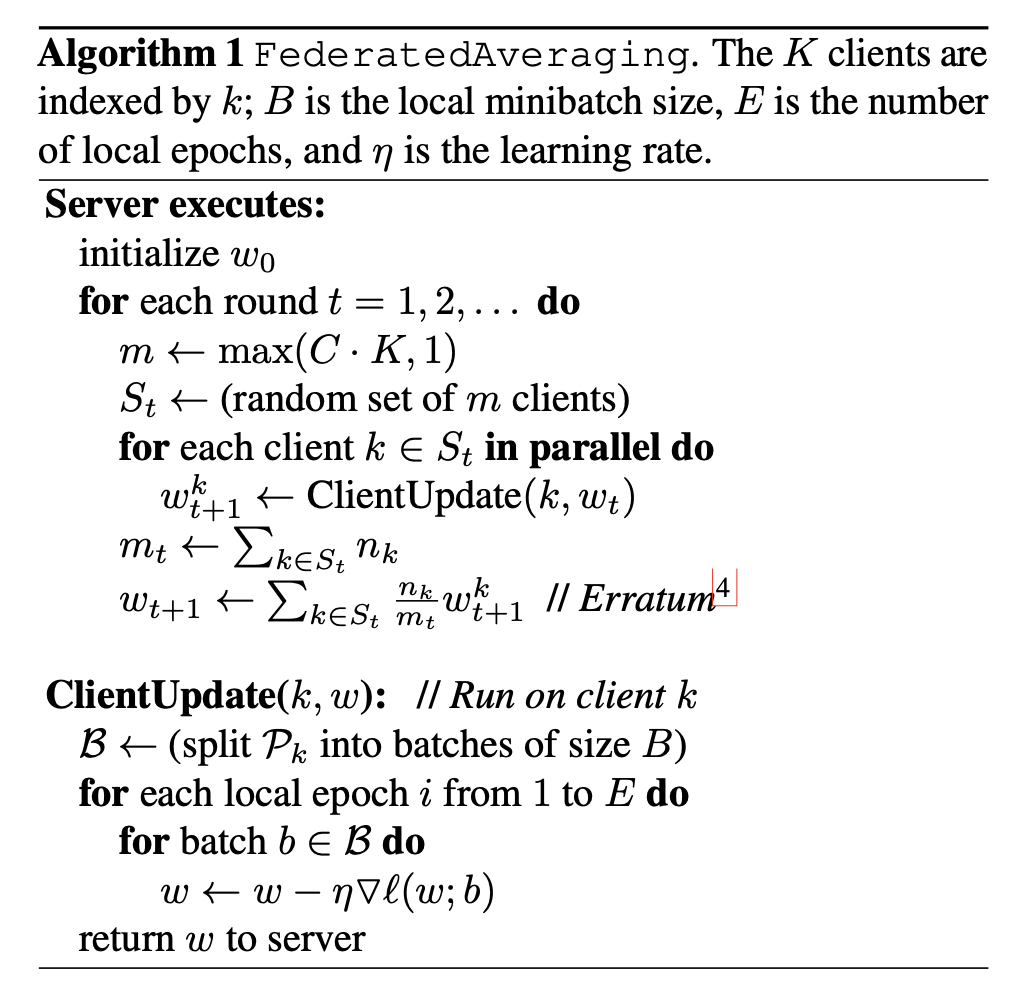}
    \label{fig:sub2}
  \end{minipage}
  \caption{a) Workflow diagram (source:\cite{educative_fedavg}) of FedAvg training pipeline. b) Training pseudo-code of FedAvg algorithm \cite{mcmahan2017communication}}
  \label{fig:both}
\end{figure}

The model weights across clients are aggregated using a weighted sum, where the weight for each client is defined based on the ratio of training examples for that client to the total global size of the dataset. The local client weights are updated using Stochastic Gradient Descent for each round of training. The algorithm has the following key hyper-parameters:\\

\begin{enumerate}
    \item \textbf{Number of Clients(C)}: The number of clients that participate in each round of training. Most of the times, a fraction of the total number of clients participate in each round.
    \item \textbf{Number of local Epochs (E):} This is the number of training passes each client makes over its local dataset on each round of training.
    \item \textbf{Local minibatch size (B):} This is the batch size for the local client updates. If the local data sizes are small, the batch size could be equivalent to the size of client dataset. In this case, if local epoch E=1, then the algorithm becomes equivalent to FedSGD. We discuss about the difference with FedAvg below.
    \item \textbf{Number of Rounds:} The number of aggregation rounds of the local client models.\\
\end{enumerate}

\textbf{FedAvg vs FedSGD}
\newline

The authors of this work make the distinction between a proposed
baseline called FedSGD and the FedAvg algorithm. In FedSGD, each
client does one single gradient update and shares the gradients
of the parameters for weighted averaging. Whereas, in FedAvg,
each client can do multiple gradient updates controlled by either
the minibatch size (B) of the number of local epochs (E) before
aggregating the actual model parameters instead of the gradients.
FedSGD has better convergence guarantees than FedAvg especially
for non convex objectives, whereas the motivation for FedAvg is
to reduce the communication overhead which is a major bottleneck
in deploying federated learning systems.\\

For general non-convex objectives, averaging models directly in parameter space could produce an arbitrarily bad model. However, in practise the authors note that FedAvg performs surprisingly well for over-parametrized models such as Neural Networks when all client models have the same random initialization. Thus, despite not having strong theoretical guarantees FedAvg is a strong contender for practical deployment of federated learning systems. \\

\section{Ranking using CTR prediction:}
\label{sec:4.3}
\vspace{\baselineskip}

\textbf{Motivation:}
\newline

 Click-through-rate prediction models which aim to predict the probability of a user clicking on an ad or an item are a critical part of online advertising and recommendation system methods. These methods can be used to predict the relevance score of each item based on the output probability and are termed as point wise ranking methods. Other methods for ranking items include pairwise and group ranking techniques which ingest a pair or set of item to calculate their relative ranking positions. 

Our motivation for choosing a CTR model for ranking was to take advantage of their ability to work with sparse and high dimensional input features. The mobile context data generated from user mobile devices and item metadata consists of lots of one-hot and multi-hot categorical features whose domain can range from thousands of possible input values. CTR models are designed to handle that level of input sparsity and can model higher order interaction across different input features. \\

\textbf{Literature Survey}
\newline

We survey popular methods for CTR prediction on the MovieLens
dataset. Logistic Regression is frequently used as the first baseline in this category, which models only the linear
combination of raw features. Google introduced the
Wide\&Deep \cite{cheng2016wide} learning system, which combines the advantages of both the linear shallow models and deep models for this task. Factorization machine models introduced in \cite{rendle2010factorization} in 2010 remain a highly competitive baseline in this task, with numerous improvements coming with deep learning models like DeepFM \cite{guo2017deepfm}. Deep Interest Network (DIN) \cite{zhou2018deep} is another CTR method proposing to use a adaptive representation vector after the embedding layer instead of a fixed dim vector to improve the expressive capability of the model. Finally, the method we implement in our federated learning CTR model pipeline is the best performing model on the MovieLens 1M benchmark called AutoInt proposed in \cite{song2019autoint}. This method utilizes multi-head self attention on top of the input embedding layers to predict a probability relevance score. We briefly describe the architecture of the model below.\\

\textbf{AutoInt CTR Model}
\newline

\begin{figure}[h]
    \centering
    \includegraphics[width=0.97\textwidth]{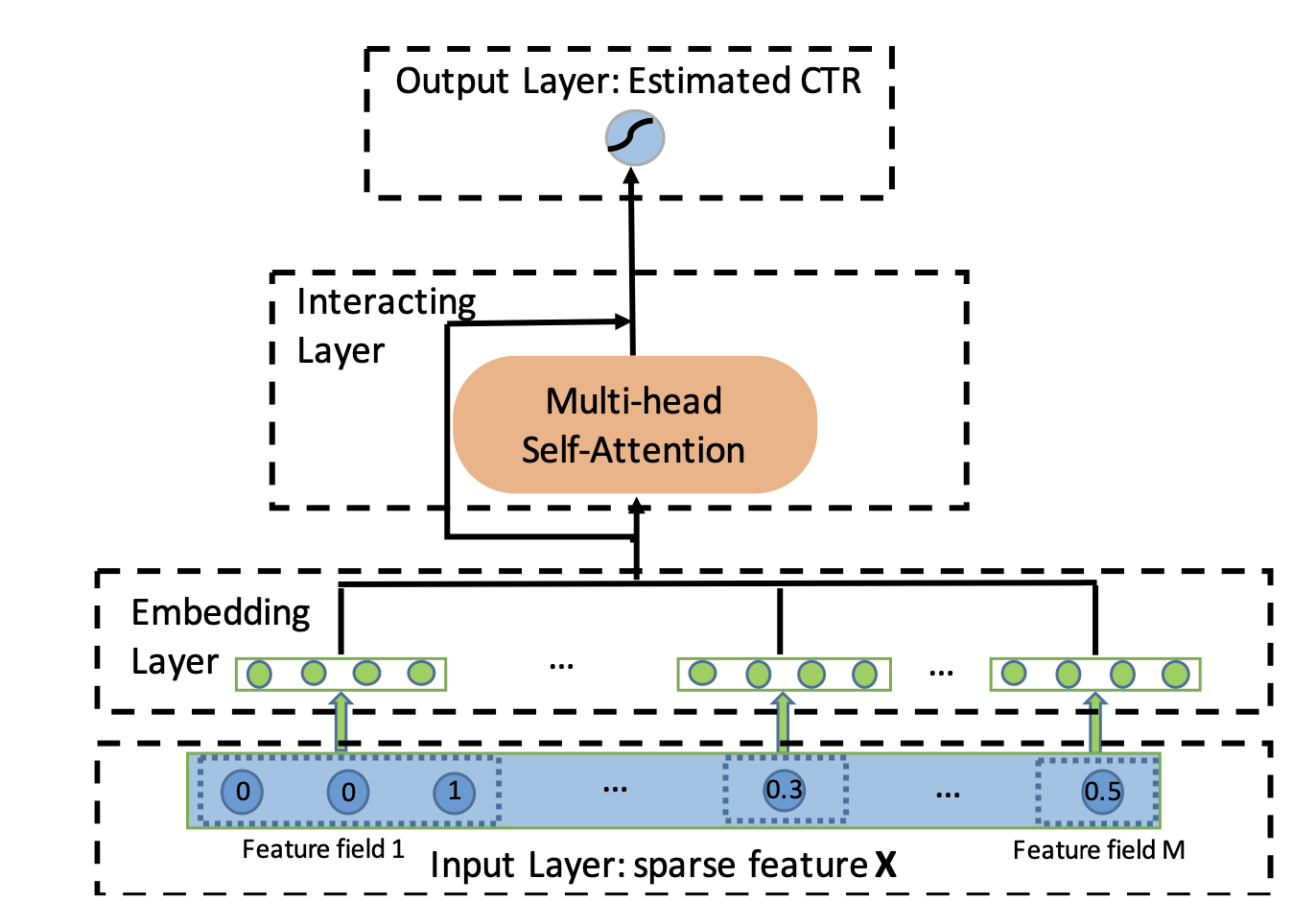}
    \caption{Model architecture of the AutoInt CTR model from Fig.1 \cite{song2019autoint}}
    \label{fig:4.2}
    \vspace{5mm}
\end{figure}

The AutoInt model architecture consists of 3 modules consisting of an embedding layer, multi-head self-attention transformer layers, and a final MLP layer that outputs sigmoid probabilities. The embedding layer accepts all type of input features: categorical, numerical and multi-valued categorical features and transforms them into a fixed dimension embedding vector. The MovieLens dataset for example consists of genre as a multi-valued feature, user occupation, zip code, movie id etc. as categorical features and age, timestamp as numerical features. \\

\vspace{10mm}
For numerical features, the embedding vector is obtained by multiplying the scalar numerical value with a single embedding vector defined for its field m.
$$e_m = \mathbf{v_m} * x_m,$$

where $\mathbf{v_m}$ is the embedding vector for $m^{th}$ numerical feature field and $x_m$ is the scalar value.  \\

For categorical features, each unique value of the feature denotes its own embedding vector:
$$e_i = \mathbf{V_i}x_i, $$

where $Vi$ is an embedding matrix for field $i$, and $x_i$ is a one-hot vector. Embedding vectors for multi-valued features are computed using the average of the embedding vectors corresponding to each value as:

$$e_i = \mathbf{V_i} x_i /q , $$

where $q$ denotes the total the number of values in that sample for the $i^{th}$ multi-valued feature and $x_i$ is the multi-hot vector representation for this feature.\\

For the model architecture, we use the same number of layers and embedding size as discussed in the original AutoInt work.  We set the embedding dimension to 16 and the number of hidden units in the last layer to 32 before the sigmoid output. Within the attention layer, we use two heads in the multi-head transformer with layer size of 32. We use the dropout value of 0.3 to prevent overfitting.

\section{Experiments}

\subsection{Setup}
\label{sub:4.4.1}

\textbf{MovieLens 1M dataset:}
\newline

We use the same MovieLens 1M \cite{movielens_dataset} dataset used in the previous centralized stage \ref{sec:4.4} to evaluate our federated learning pipeline. The motivation for choosing this dataset was availability of public benchmarks such as the AutoInt \cite{song2019autoint} for the CTR prediction task. Further, in absence of the company’s private dataset, this dataset was a good alternative as it contained both user features and item features which correspond to the real-world data in our pipeline. 

We follow the same methodology as AutoInt \cite{song2019autoint} for transforming the MovieLens dataset for a CTR prediction task. The ratings ranging from [1-5] are transformed into binary “relevant/non-relevant” corresponding to “clicked/not clicked” notion of CTR prediction task by thresholding all ratings above 3 as relevant. Ratings less than 3 are treated as non relevant samples and equal to 3 are discarded. The total size of the dataset after this pre-processing step is $\approx 730k$ ratings. \\


\textbf{Evaluation Metrics}

We use two popular metrics to evaluate our experiments:\\

\begin{enumerate}
    \item \textbf{Area Under the ROC Curve (AUC):} It measures the probability that a CTR predictor will assign a higher score to a randomly chosen positive item than a randomly chosen negative item. A higher AUC indicates a better performance. It is a measure of aggregate performance across different classification thresholds.  We evaluate it at 10 thresholds divided equally between 0 to 1.

    In federated learning setting, the AUC is calculated individually for each client test set and then aggregated using weighted sum. The weights are calculated the same way as calculated in FedAvg, by using the fraction of client dataset size to the global dataset size as below.

    $$\text{AUC}_{\text{federated}} = \sum_{i=1}^{N} \frac{|D_i|}{|D|} \times \text{AUC}_{\text{client}}(D_i)$$
    
    \item \textbf{Log Loss:} Since all models attempt to minimize the binary cross entropy log loss defined below, we use it as a straightforward metric. Lower log loss indicates better performance. It is to noted that a slightly higher AUC or lower Log loss at 0.001-level is regarded significant for CTR prediction task, as pointed out in previous literature \cite{cheng2016wide}\cite{song2019autoint}.
    $$\text { Logloss }=-\frac{1}{N} \sum_{j=1}^N\left(y_j \log \left(\hat{y}_j\right)+\left(1-y_j\right) \log \left(1-\hat{y}_j\right)\right)$$ 

    For federated learning setting, the log loss is calculated in a similar weighted sum fashion as AUC above. \\
\end{enumerate}

\subsection{Federated vs Centralized Evaluation:}

We conduct experiments to compare performance of the company’s previously deployed Logistic Regression model against the AutoInt model. Logistic Regression remains a good baseline in CTR predictions task to understand the efficacy of higher order combination of input features in the task as opposed to linear combination. For Logistic Regression model, we experiment with the following two methods for the input layer:\\

\begin{enumerate}
    \item \textbf{Raw vectors:} Directly feeding raw vectors comprised of numerical, categorical and multi-valued features. Numerical features are min-max normalized in the range [0-1], whereas categorical and multi-valued features are one-hot and multi-hot encoded in boolean 1s/0s respectively. This was the company’s current approach to compute ranking scores using this model. This method corresponds to the LR + Raw Inputs model in the experiment \textbf{Table \ref{tab:4.1}}. \\
    \item \textbf{Embedding vectors:} In this approach, we add an embedding layer before the Logistic Regression model to transform the raw inputs into fixed low-dimensional embedding vectors of size 16. We follow the same methodology discussed for the AutoInt model in \textbf{Section \ref{sec:4.3}} to compute these embedding vectors using the embedding table. The embedding vector for each feature is concatenated forming a 7 (no. of input features ) x 16(embedding dim) = 112 dimensional input. This method corresponds to the LR + Embedding input experiment in \textbf{Table \ref{tab:4.1}}.\\
\end{enumerate}

\textbf{Centralized vs Federated Experiments}
\newline

The authors of the AutoInt \cite{song2019autoint} paper discuss the performance of this CTR prediction model in a centralized machine learning setting where the entire dataset resides in one place. In our two-stage setting however, we want to deploy these CTR models on user devices and update the model weights using the FedAvg algorithm. Therefore, we conduct experiments first in a centralized setting similar to the paper, to verify and compare results with the above two Logistic Regression methods. And then further, we conduct the same experiments in federated learning setting by splitting the dataset across 10 clients using both IID and non-IID splits. We describe our experiments below in more detail.\\

\begin{table}[h]
\begin{tabular}{|c|cc|cccc|}
\hline
\rowcolor[HTML]{EFE8A0}
\textbf{Setting}     & \multicolumn{2}{c|}{\cellcolor[HTML]{EFE8A0}\textbf{Centralized}} & \multicolumn{4}{c|}{\cellcolor[HTML]{EFE8A0}\textbf{Federated}}                                                                    \\ \hline
\textbf{Split}       & \multicolumn{2}{c|}{\textbf{}}                                    & \multicolumn{2}{c|}{\textbf{IID}}                                          & \multicolumn{2}{c|}{\textbf{Non-IID}}                 \\ \hline
\textbf{Model}       & \multicolumn{1}{c|}{\textbf{AUC}}       & \textbf{Log Loss}       & \multicolumn{1}{c|}{\textbf{AUC}} & \multicolumn{1}{c|}{\textbf{Log Loss}} & \multicolumn{1}{c|}{\textbf{AUC}} & \textbf{Log Loss} \\ \hline
LR + Raw Inputs      & 0.751                                   & 0.455                   & 0.692                             & \multicolumn{1}{c|}{0.514}             & 0.671                             & 0.527             \\
LR + Embedding Input & 0.770                                   & 0.442                   & 0.766                             & \multicolumn{1}{c|}{0.445}             & 0.741                             & 0.454             \\
AutoInt              & 0.848                                   & 0.367                   & 0.796                             & \multicolumn{1}{c|}{0.425}             & 0.772                             & 0.434             \\ \hline
\end{tabular}
\caption{Summary table of experiments across centralized and federated learning setting. The federated experiments are further divided based on the dataset split strategy across clients.}
\label{tab:4.1}
\end{table}

\textbf{Split Strategy: IID vs Non-IID splits}
\newline

A key challenge in deploying federated learning systems is that client data distributions are not representative of the population distributions as a whole. Some mobile users will make much heavier use of the service or app than others, or a set of users may have entirely different preferences or mobile activity habits. Therefore the client data distributions are not “Independent and identically distributed” or are non-IID sampled. We therefore, conduct our federated learning experiments considering both IID and non-IID splits.\\

\begin{itemize}
    \item \textbf{IID Split:} We obtain IID split of MovieLens data by uniform random sampling across 10 equal size splits. After the pre-processing step of binarization of ratings described in \textbf{Section \ref{sub:4.4.1}}, the MovieLens 1M dataset consists of $\approx 730k$ ratings which are then divided into train, val, test splits in the ratio of 80-10-10. Each of the splits is then divided with uniform random sampling into 10 client datasets. Each of the client dataset consists of $\approx 58k$ ratings for training set and $\approx 7.1k$ ratings for the test set.\\
    
    \item \textbf{Non-IID splits:} There are three key approaches used in federated learning literature to divide a dataset into non-IID splits. For classification problems, labels are unevenly divided between equal size client datasets as suggested in \cite{mcmahan2017communication}.  Another approach is to use Drichlet distribution to sample different size client datasets ignoring the label distribution. This introduces heterogeneity in the number of weight updates since client mini-batch sizes are kept consistent. Another approach is to divide the dataset based on patterns in feature space using clustering algorithms.
    
    The last approach is the most realistic simulation of our use case where there would be different clusters of users with similar mobile usage behaviour. We therefore use this strategy to split the dataset into 10 clusters using k-NN clustering algorithm. We obtain embeddings for each user in the dataset using SVD decomposition of the user-item interaction matrix. This embedding obtained for each user is a 50-dim vector based on 50 singular values. We then apply kNN-clustering  on these user embeddings. The cluster number is used to split the dataset into corresponding clients. A distribution of the train and test client test datasets is plotted below.\\
\end{itemize}

\begin{figure}[h]
    \centering
    \includegraphics[width=0.85\textwidth]{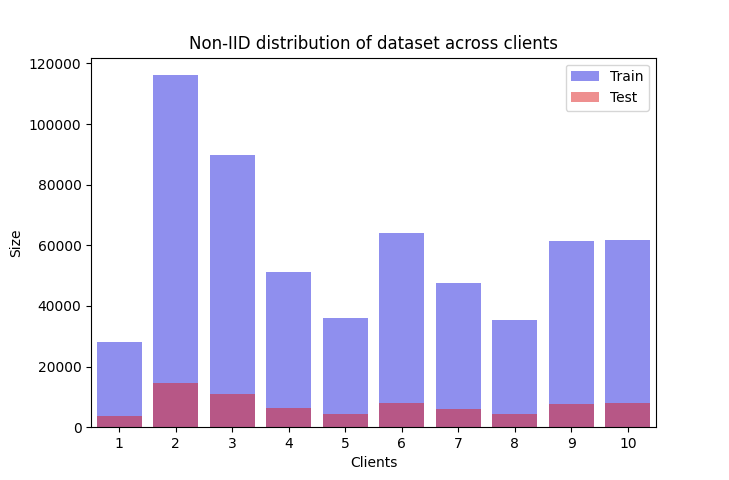}
    \caption{Non-IID Distribution of MovieLens 1M dataset across 10 clients.}
    \label{fig:4.3}
\end{figure}

\subsection{Ablation Experiments of the Federated Model:}

A key consideration of deploying a Federated learning pipeline is the bandwidth cost involved in the sharing of model weights from the edge devices to the central aggregation server. As opposed to central optimization where computational costs dominate, communication costs are a key parameter of Federated optimization. Therefore, we wanted to investigate if we need to federate all parts of our model architecture or we can get away by keeping some model parameters completely local to client devices. Our hypothesis was that in our actual real world pipeline, our client datasets will be non-IID, and thus it might be better to have some non-shared parameters which can be used to personalize to each client dataset. We thus use the non-IID split of the MovieLens 1M dataset defined previously, for this experiment.\\

\textbf{Experiment Summary}
\newline

We conduct architecture ablation experiments where for each of the 3 parts of the AutoInt CTR model architecture, we only federate one part at a time. This means that the model parameters are aggregated globally across clients only for that part of the model, whereas the other parameters remain local.  The 3 parts involved in the architecture described in Figure \ref{fig:4.2} corresponding to our experiments are: \\

\begin{itemize}
    \item \textbf{Embedding layer:} In this experiment, the first or the embedding layer weights are shared across clients. The rest of the two layers are kept local. The total number of parameters corresponding to this layer are: 57600.
    \item \textbf{Attention or the Interaction layer:} Similarly, the multi-head attention weights are federated across clients. The total number of parameters corresponding to this layer are: 38015.
    \item \textbf{Output layer:} This layer consists of a simply a downsampling layer followed by a sigmoid activation that outputs a probability CTR score. The total number of parameters corresponding to this layer are: 441.\\
\end{itemize}

Since this was not natively supported by the Flower federated learning framework, we implement functionality in the library to selectively store parts of client weights as storage dumps, and federate the rest of the parameters. In the next round of training, the non-federated weights are loaded from the local client storage. We detail the results obtained for these experiments below.\\

We show the below results after 20 rounds of training.\\

\begin{table}[h]
\begin{tabular}{|c|c|c|}
\hline
\rowcolor[HTML]{EFE8A0} 
\textbf{Experiments using Non-IID split} & \textbf{Test AUC} & \textbf{Test Log Loss} \\ \hline
All layers federated                     & 0.7728            & 0.434                  \\ \hline
\textbf{Embedding layer Federated}       & \textbf{0.7971}   & \textbf{0.423}         \\ \hline
Attention layer Federated                & 0.7878            & 0.491                  \\ \hline
Output layer Federated                   & 0.7852            & 0.504                  \\ \hline
\end{tabular}
\caption{Results table for Ablation experiments on the AutoInt model architecture}
\label{tab:ablation}
\end{table}

\begin{figure}[h]
  \centering
  \begin{minipage}[b]{0.49\linewidth}
    \includegraphics[width=\linewidth]{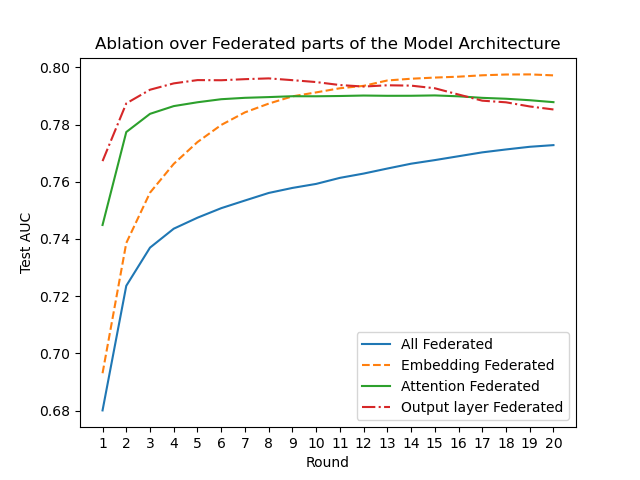}
  \end{minipage}
  \hfill
  \begin{minipage}[b]{0.49\linewidth}
    \includegraphics[width=\linewidth]{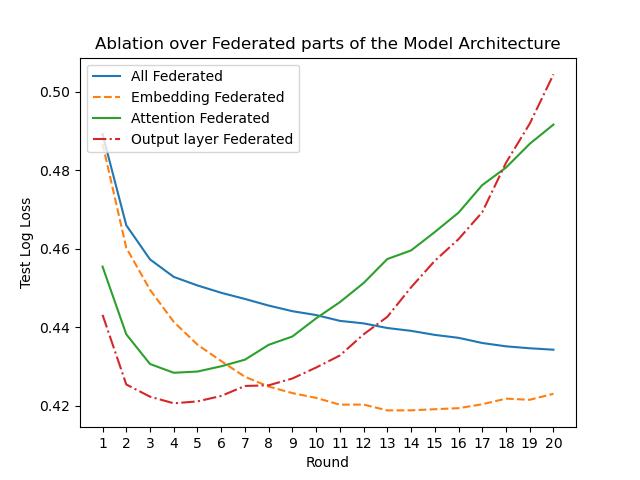}
  \end{minipage}
  \caption{Test AUC and LogLoss plots across federated aggregation rounds for Ablation experiments}
  \label{fig:ablation_plots}
\end{figure}

As observed in the experiment results, federating only the Embedding layer and keeping other parts local performs better than the default All federated setting. For other experiments, such as Attention and Output layer, the Test log loss diverges after a few rounds of training. Note that both Test metrics: AUC and Log loss are obtained by weighted averaging of individual client test datasets. We discuss in detail the results of this and the previous experiment in the Results and conclusion section below.

\section{Results Discussion \& Conclusion:}

\vspace{5mm}
\textbf{Centralized vs Federated Evaluation Experiment}
\newline

Based on the experiments detailed in \textbf{Table \ref{tab:4.1}}, we draw the following conclusions on the obtained results:\\

\begin{enumerate}
    \item The AutoInt CTR prediction model performs consistently better with an almost $\approx10\%$ improvement in AUC over the company’s previous Logistic Regression baseline across both centralized and federated evaluation settings. We were able to reproduce the results cited in Table-2 of the original paper \cite{song2019autoint} regarding the model’s performance in the centralized setting.\\
    \item There is a consistent decline of performance when the models are trained in federated setting as compared to the centralized setting. This is observed across all 3 model categories. This is expected as averaging of models in parameter space can lead to arbitrarily bad models especially for non-convex objectives observed with neural network models.\\
    \item Adding an embedding layer that transforms raw one-hot feature vectors into learnable low-dim continuous embedding vectors leads to significant improvement of AUC in even simple models like logistic regression. The AUC improved from 0.75 to 0.77 in centralized setting and even more so in federated setting from 0.69 to 0.76 in IID split.\\
    \item It is interesting to note that in case of LR + Embedding layer experiment, the decline of model performance when federating the model is fairly small from 0.77 in centralized setting to 0.76 in IID split and 0.741 in case of non-IID split respectively. This is in contrast to the AutoInt where a relatively sharper decline from 0.84 to 0.79 is observed between the centralized and federating IID setting.\\
    \item Overall, the AutoInt model performs best across all settings as compared to the company’s previous baseline of using Logistic Regression as the main model for ranking items in the second stage. Thus it becomes the new model that we implemented in Kotlin Multiplatform for deployment in mobile devices for the Federated stage of our pipeline.\\
\end{enumerate}

\textbf{Ablation Experiments over the Federated Model architecture}\\

Based on the experiment results obtained in \textbf{Table \ref{tab:ablation}}, we draw the following conclusions:\\

\begin{enumerate}
    \item The best results with a Test AUC of 0.797 were obtained when only the Embedding layer containing the most parameters (num parameters = 57600) was federated and the other two layer parameters were kept local for each client. This performed even better than the previous best Test AUC of 0.772 where the entire model is federated for the non-IID split experiment \textbf{Table \ref{tab:4.1}}. We hypothesize that this could be because that the other layer parameters containing the Attention layer and the output layer can model interactions specific to the local client datasets and averaging those parameters globally leads to a decline in performance.\\
    
    \item  Federating the first layer or the embedding layer leads to developing a shared vocabulary between client models. Federated any other following part of the model architecture such as the Attention, output layer doesn’t have the same effect and leads to divergence of Log loss as number of rounds progresses. This is evident in results obtained for the experiments where only the Attention or the Output layer are federated. \\
\end{enumerate}

Overall, In this chapter, we explained our methodology for implementing the Federated ranking stage of our pipeline using the AutoInt model architecture. We conducted experiments for both centralized and federated evaluation and compared results against the previous Logistic Regression model. Although extensive further experiments are needed to evidently conclude our ablation study results, we observed some a direction by which we can shave off some of client’s communication bandwidth by federating only the first Embedding layer of the model. This can be thoroughly concluded as part of future work.

\chapter{Conclusion}

\section{Key Contributions}

We summarize the key contributions of the work done during this internship below.\\

\begin{itemize}
    \item Breaking down the initial problem statement into clear parts of: “when” to recommend content and “what” is the best content to show. We isolate and tackle the latter problem of what content to show during this internship.\\
    \item Since the company’s existing Federated learning pipeline was entered around a simple Logistic Regression model to keep the computational cost of edge device training low, we proposed ways to iterate over better input representation rather than model through the Activity recognition task. This led to a conclusive usage of session based feature embeddings as input for our downstream pipeline. We experimented in Python and finally implemented from scratch the proposed solution in Kotlin Multiplatform language for end-to-end deployment.\\
    \item A key contribution of this internship was the proposal of the Two-stage recommendation pipeline with the first stage being a Centralized stage which can be hot-swapped with an App’s existing recommendation system. The second stage is the Federated stage which ranks items based on mobile context. The previous methodology was based on a single Federated Logistic Regression model that finds top-k content directly using all data sources.\\
    \item The final output of the internship was a Machine Learning library for Kotlin Multiplatform where plug-and-play modules such as multi-head attention, MLP, Embedding layers etc. were implemented. The forward, backward passes of these modules were written from scratch in Kotlin as no existing ML library that supports auto-differentiation in Kotlin was found in our survey that supports both Android/iOS devices.
\end{itemize}

\section{Challenges}

\begin{itemize}
    \item A major challenge during this internship was the absence of a Recommendation dataset which contained mobile sensor data similar to our problem statement. We surveyed multiple public datasets, but we didn't find any direct match with our statement.\\
    \item Absence of a private company dataset for the end-to-end testing of our proposed pipeline on live user devices was also a challenge. We therefore chose the closest fit of MovieLens 1M dataset which we used for evaluation of both stages of our pipeline.\\
    \item Absence of a mature ML eco-system in Kotlin language meant that we had to write a lot of boiler plate code which is generally available as import from libraries. This included evaluation metrics, Fast Fourier transform, a mature ML library that supports autodiff and quick implementation of custom models. Thus, significant engineering efforts were a part of this internship which we considered as an opportunity for growth.
\end{itemize}

\section{Future work}

We propose that following future work can concretize the results and conclusions made in this work. \\

\begin{itemize}
    \item An end-to-end testing of the complete pipeline with live Federated training on mobile devices is a key area for future work in this project.\\
    \item Further experiments for the Federated stage pipeline, including generalization of the observed results for a large number of clients, and continuous rounds of training.\\
    \item Benchmarks quantifying the wall clock time of training on different edge mobile devices. This would be helpful to determine a trade-off between better models vs the training/inference time.
\end{itemize}

\bibliographystyle{plainurl}     
\def\bibname{References}
\bibliography{ref}     

\end{document}